\documentclass[11pt]{article}
\usepackage{amsmath,amssymb,amsthm}

\usepackage{enumerate,subfigure,multirow,epsfig,psfrag,hhline,color,srcltx,subeqnarray}
\usepackage{tabularx}
\usepackage{tabulary}
\usepackage{xcolor}
\usepackage{hyperref}
\DeclareGraphicsExtensions{.pdf,.png,.jpg}
\usepackage[sort]{cite}
\usepackage[sort&compress,square,comma,authoryear]{natbib}

\usepackage{tcolorbox}
\hypersetup{breaklinks=true,colorlinks=true,linkcolor=blue,citecolor=blue}
\usepackage{appendix}
\usepackage{graphicx}
\usepackage{graphics}

\usepackage{algorithm}
\usepackage{algpseudocode}
\usepackage[noabbrev,capitalise,nameinlink]{cleveref}
\usepackage{float}
\floatstyle{boxed}
\restylefloat{figure}

\newtheorem{assumption}{Assumption}
\newtheorem{remark}{Remark}
\newtheorem{lemma}{Lemma}

\newtheorem{proposition}{Proposition}
\newtheorem{theorem}{Theorem}

\newtheorem{definition}{Definition}

\DeclareMathOperator*{\argmin}{arg\,min}

\allowdisplaybreaks

\title{Target-Based Temporal-Difference Learning}

\author{Donghwan Lee\thanks{D. Lee is with Coordinated Science Laboratory (CSL),
University of Illinois at Urbana-Champaign, IL 61801, USA {\tt\small
donghwan@illinois.edu}.}  and Niao He
\thanks{N. He is with the Department of Industrial and Enterprise Systems Engineering,
University of Illinois, Urbana-Champaign, IL 61801, USA {\tt\small
niaohe@illinois.edu}.} }

\begin{document}

\maketitle

\begin{abstract}
The use of target networks has been a popular and key component of recent deep Q-learning algorithms for reinforcement learning, yet little is known from the theory side. In this work, we introduce a new family of target-based temporal difference (TD) learning algorithms and provide theoretical analysis on their convergences. In contrast to the standard TD-learning, target-based TD algorithms maintain two separate learning parameters--the target variable and online variable. Particularly, we introduce three members in the family, called the {\em averaging TD, double TD}, and {\em periodic TD}, where the target variable is updated through an averaging, symmetric, or periodic fashion, mirroring those techniques used in deep Q-learning practice.

We establish asymptotic convergence analyses for both {\em averaging TD} and {\em double TD} and a finite sample analysis for {\em periodic TD}. In addition, we also provide some simulation results showing potentially superior convergence of these target-based TD algorithms compared to the standard TD-learning. While this work focuses on linear function approximation and policy evaluation setting, we consider this as a meaningful step towards the theoretical understanding of deep Q-learning variants with target networks.
\end{abstract}

\section{Introduction}
Deep Q-learning~\citep{mnih2015human} has recently captured significant attentions in the reinforcement learning (RL) community for outperforming human in several challenging tasks. Besides the effective use of deep neural networks as function approximators, the success of deep Q-learning is also indispensable to the utilization of a separate target network for calculating target values at each iteration. In practice, using target networks is proven to substantially improve the performance of Q-learning algorithms, and is gradually adopted as a standard technique in modern implementations of Q-learning.

To be more specific, the update of Q-learning with target network can be viewed as follows:
\begin{align*}
&\theta_{t+1}=\theta_t+\alpha(y_t-Q(s_t,a_t;\theta_t))\nabla_\theta Q(s_t,a_t;\theta_t)
\end{align*}
where $y_t=r(s_t,a_t)+\gamma\max_a Q(s_{t+1},a;\theta'_t)$, $\theta_t$ is the online variable, and $\theta'_t$ is the target variable. Here the state-action value function $Q(s,a;\theta)$ is parameterized by $\theta$. The update of the {\rm online variable} $\theta_t$ resembles the stochastic gradient descent step. The term $r(s_t,a_t)$ stands for the intermediate reward of taking action $a_t$ in state $s_t$, and $y_t$ stands for the target value under the {\rm target variable}, $\theta'_t$. When the target variable is set to be the same as the online variable at each iteration, this reduces to the standard Q-learning algorithm~\citep{watkins1992q}, and is known to be unstable with nonlinear function approximations. Several choices of target networks are proposed in the literature to overcome such instability: (i) periodic update, i.e., the target variable is copied from the online variable every $\tau >0$ steps, as used for deep Q-learning~\citep{mnih2015human,wang2016dueling,mnih2016asynchronous,gu2016continuous}; (ii) symmetric update, i.e., the target variable is updated symetrically as the online variable; this is first introduced in double Q-learning~\citep{hasselt2010double,van2016deep}; and (iii) Polyak averaging update, i.e., the target variable takes weighted average over the past values of the online variable; this is used in deep deterministic policy gradient~\citep{lillicrap2015continuous,heess2015memory} as an example. In the following, we simply refer these as target-based Q-learning algorithms.

While the integration of Q-learning with target networks turns out to be successful in practice, its theoretical convergence analysis remains largely an open yet challenging question. As an intermediate step towards the answer, in this work, we first study target-based temporal difference (TD) learning algorithms and establish their convergence analysis. TD algorithms~\citep{sutton1988learning,sutton2009fast,sutton2009convergent} are designed to evaluate a given policy and are the fundamental building blocks of many RL algorithms. Comprehensive surveys and comparisons among TD-based policy evaluation algorithms can be found in~\cite{dann2014policy}. Motivated by the target-based Q-learning algorithms~\citep{mnih2015human,wang2016dueling}, we introduce a target variable into the TD framework and develop a family of target-based TD algorithms with different updating rules for the target variable. In particular, we propose three members in the family, the {\em averaging TD, double TD}, and {\em periodic TD}, where the target variable is updated through an averaging, symmetric or periodic fashion, respectively. Meanwhile, similar to the standard TD-learning, the online variable takes stochastic gradient steps of the Bellman residual loss function while freezing the target variable. As the target variable changes slowly compared to the online variable, target-based TD algorithms are prone to improve the stability of learning especially if large neural networks are used, although this work will focus mainly on TD with linear function approximators.

Theoretically, we prove the asymptotic convergence of both {\em averaging TD} and {\em double TD}. We also provide a finite sample analysis for the {\em periodic TD} algorithm. Practically, we also run some simulations showing superior convergence of the proposed target-based TD algorithms compared to the standard TD-learning. In particular, our empirical case studies demonstrate that the target TD-learning algorithms  outperforms the standard TD-learning in the long run with smaller errors and lower variances,  despite their slower convergence at the very beginning. Moreover, our analysis reveals an important connection between the TD-learning and the target-based TD-learning. We consider the work as a meaningful step towards the theoretical understanding of deep Q-learning with general nonlinear function approximation.

\textbf{Related work.}  The first target-based reinforcement learning was proposed in~\citep{mnih2015human} for policy optimization problems with nonlinear function approximation, where only empirical results were given. To our best knowledge, target-based reinforcement learning for policy evaluation has not been specifically studied before. A somewhat related family of algorithms are the gradient TD (GTD) learning\textbf{} algorithms~\citep{sutton2009fast,sutton2009convergent,mahadevan2014proximal,dai2017learning}, which minimize the projected Bellman residual through the primal-dual algorithms. The GTD algorithms share some similarities with the proposed target-based TD-learning algorithms in that they also maintain two separate variables -- the primal and dual variables, to minimize the objective. Apart from this connection, the GTD algorithms are fundamentally different from the averaging TD and double TD algorithms that we propose. The proposed periodic TD algorithm can be viewed as approximately solving least squares problems across cycles, making it closely related to two families of algorithms,  the least-square TD (LSTD) learning algorithms~\citep{bertsekas1995dynamic,Bradtke1996} and the least squares policy evaluation (LSPE)~\citep{bertsekas2009projected,yu2009convergence}. But they also distinct from each other in terms of the subproblems and subroutines used in the algorithms. Particularly,  the periodic TD executes stochastic gradient descent steps while LSTD uses the least-square parameter estimation method to minimize the projected Bellman residual. On the other hand, LSPE directly solves the subproblems without successive projected Bellman operator iterations. Moreover, the proposed periodic TD algorithm enjoys a simple finite-sample analysis based on existing results on stochastic approximation.


\section{Preliminaries}\label{sec:preliminaries}
In this section, we  briefly review the basics of the TD-learning algorithm with linear function
approximation. We first list a few notations that will be used throughout the paper.

\paragraph{Notation} The following notation is adopted: for a convex
closed set $\cal S$, $\Pi_{\cal S}(x)$ is the projection of $x$
onto the set $\cal S$, i.e., $\Pi_{\cal S}(x):={\rm argmin}_{y\in {\cal
S}} \|x-y\|_2$; ${\rm diam}({\cal S}):=\sup_{x \in {\cal S},y \in {\cal
S}}\|x-y\|_2 $ is the diameter of the set $\cal S$; $\|x\|_D:=\sqrt{x^T Dx}$ for any
positive-definite $D$; $\lambda_{\min}(A)$ and $\lambda_{\max}(A)$ denotes the minimum and maximum eigenvalues of a symmetric matrix $A$, respectively.

\subsection{Markov Decision Process (MDP)}\label{subsection:MDP}
In general, a (discounted) Markov
decision process is characterized by the tuple ${\cal M}: =
({\cal S},{\cal A},P,r,\gamma)$, where ${\cal S}$ is a finite
state space, $\cal A$ is a finite action
space, $P(s,a,s'):={\mathbb P}[s'|s,a]$ represents the (unknown)
state transition probability from state $s$ to $s'$ given action
$a$, $r:{\cal S}\times {\cal A}\to
[0,\sigma]$ is a uniformly bounded stochastic reward, and $\gamma \in (0,1)$ is the discount factor. If action
$a$ is selected with the current state $s$, then the state
transits to $s'$ with probability $P(s,a,s')$ and incurs a random
reward $r(s,a) \in [0,\sigma ]$ with expectation
$R(s,a)$. A stochastic policy is a distribution $\pi \in \Delta_{|{\cal S}|\times |{\cal A}|}$ representing the probability $\pi(s,a)={\mathbb
P}[a|s]$, $P^\pi$ denotes the transition matrix whose $(s,s')$
entry is ${\mathbb P}[s'|s] = \sum_{a \in {\cal A}} {P(s,a,s')\pi
(s,a)}$, and $d \in \Delta_{|{\cal S}|}$ denotes the stationary
distribution of the state $s\in {\cal S}$ under policy $\pi$, i.e., $d = d P^\pi$. The following assumption is standard in the literature.
\begin{assumption}\label{assumption:positive-stationary-dist}
We assume that $d(s)>0$ for all $s\in {\cal S}$.
\end{assumption}

We also define
$r^\pi(s)$ and $R^\pi(s)$ as the stochastic reward and its expectation given the policy $\pi$ and the
current state $s$, i.e.
\begin{align*}
R^\pi(s)&:=\sum_{a\in {\cal A}}{\pi (s,a)R(s,a)}.
\end{align*}
The infinite-horizon discounted value function given policy $\pi$ is
\begin{align*}
&J^\pi(s):={\mathbb E}\left[ {\left. \sum_{k=0}^\infty {\gamma^k r(s_k,a_k)}\right|s_0=s} \right],
\end{align*}
where $s\in {\cal S}$, ${\mathbb E}$ stands for the expectation
taken with respect to the state-action-reward trajectories.

\subsection{Linear Function Approximation}
Given pre-selected basis (or feature)
functions $\phi_1,\ldots,\phi_n:{\cal S}\to {\mathbb R}$, $\Phi
\in {\mathbb R}^{|{\cal S}| \times n}$ is defined as
\begin{align*}
\Phi:=\begin{bmatrix}
   \phi(1)^T\\
   \phi(2)^T\\
    \vdots \\
   \phi(|{\cal S}|)^T\\
\end{bmatrix} \in {\mathbb R}^{|{\cal S}| \times n}, \text{ where }
\phi(s):=\begin{bmatrix}
   \phi_1(s)\\
   \phi_2(s)\\
   \vdots \\
   \phi_n(s)\\
\end{bmatrix} \in {\mathbb R}^n.
\end{align*}
Here $n \ll |{\cal S}|$ is a positive integer and $\phi(s)$
is a feature vector. It is standard to assume that the columns of $\Phi$ do not have any redundancy up to linear combinations. We make the following assumption.
\begin{assumption}
$\Phi$ has full column rank.
\end{assumption}

\subsection{Reinforcement Learning (RL) Problem}
In this paper, the goal of RL with the linear function approximation is to find the weight vector $\theta \in {\mathbb R}^n$ such that
$J_{\theta}:=\Phi\theta$ approximates the true value function $J^{\pi}$.
This is typically done by minimizing the {\em mean-square Bellman
error} loss function~\citep{sutton2009fast}
\begin{align}
\min_{\theta\in {\mathbb R}^n} l(\theta)\nonumber
&:=\frac{1}{2}{\mathbb E}_{s} [([{\mathbb E}_{s',r} [r(s,a)+
\gamma J_\theta(s')]-J_\theta(s)])^2]\nonumber\\
&=\frac{1}{2} \|R^{\pi}+\gamma P^\pi \Phi\theta-\Phi\theta\|_{D}^2,\label{eq:loss-function1}
\end{align}
where $D$ is defined as a diagonal matrix with diagonal entries
equal to a stationary state distribution $d$ under the policy $\pi$.
Note that due to~\cref{assumption:positive-stationary-dist}, $D\succ 0$. In typical RL setting, the model is unknown, while only samples of the state-action-reward are observed. Therefore, the problem can only be solved in stochastic way using the observations. In order to formally analyze the sample complexity, we consider the following assumption on the samples.
\begin{assumption}\label{assumption:sampling-oracle}
There exists a Sampling Oracle (SO) that takes input $(s,a)$ and generates a new state $s'$ with probabilities $P(s,a,s')$ and a stochastic reward $r(s,a) \in [0,\sigma ]$.
\end{assumption}

This oracle model allows us to draw i.i.d. samples $(s,a, r, s')$ from $s\sim d(\cdot), a\sim\pi(s,\cdot), s'\sim P(s,a,\cdot)$. While such an i.i.d. assumption may not necessarily hold in practice, it is commonly adopted for complexity analysis of RL algorithms in the literature~\citep{sutton2009fast,sutton2009convergent,bhandari2018finite,dalal2018finite}. It's worth mentioning that several recent works also provide complexity analysis when only assuming Markovian noise or exponentially $\beta$-mixing properties of the samples~\citep{Antos2008,bhandari2018finite,dai18sbeed,srikant2019}. For sake of simplicity, this paper only focuses on the i.i.d. sampling case.


A naive idea for solving \ref{eq:loss-function1} is to apply the stochastic gradient descent steps,
$\theta_{k+1}  = \theta_k- \alpha_k \tilde\nabla_\theta l(\theta_k)$, where $\alpha_k>0$ is a step-size and $\tilde\nabla_\theta l(\theta_k)$ is a stochastic estimator of the true gradient of $l$ at $\theta=\theta_k$,
\begin{align*}
\nabla_\theta l(\theta_k)
 &= {\mathbb E}_{s,a}\big[({\mathbb E}_{s',r} [r(s,a) + \gamma J_{\theta_k}(s')]-J_{\theta_k}(s))^T\times ({\mathbb E}_{s'} [\gamma \nabla_\theta J_{\theta_k}(s')]-\nabla_\theta J_{\theta_k}(s))\big],
\end{align*}
 This approach is called the residual method~\citep{baird1995residual}. Its main drawback is the double sampling issue~\citep[Lemma~6.10, pp.~364]{bertsekas1996neuro}: to obtain an unbiased stochastic estimation of $\nabla_\theta l(\theta_k)$, we need two independent samples given any pair $(s,a) \in {\cal S} \times {\cal A}$. This is possible under~\cref{assumption:sampling-oracle}, but hardly implementable in most real applications.

\subsection{Standard TD-Learning}

In the standard TD-learning~\citep{sutton1988learning}, the gradient term
${\mathbb E}_{s'}[\gamma\nabla_\theta J_{\theta_k}(s')]$ in
the last line ($\nabla_\theta l(\theta_k)$) is omitted~\citep[pp.~369]{bertsekas1996neuro}. The resulting update rule is
$$\theta_{k+1}=\theta_k- \alpha_k \eta (\theta_k), \text{where}$$
$$\eta (\theta_k):=-(r(s,a)+\gamma J_{\theta_k}(s')- J_{\theta_k}(s)) \nabla_\theta J_{\theta_k}(s).
$$
While the algorithm avoids the double sampling problem and is simple to implement, a key issue here is that the stochastic gradient $\eta(\theta_k)$ does not correspond to the true gradient of the loss function $l(\theta)$ or any other objective functions, making the theoretical analysis rather subtle. Asymptotic convergence of the TD-learning was given in the original paper~\citep{sutton1988learning} in tabular case and in~\citet{tsitsiklis1997analysis} with linear function approximation. Finite-time convergence analysis was recently established in~\citet{bhandari2018finite,dalal2018finite,srikant2019}.

\paragraph{Remark.} The TD-learning can also be interpreted as minimizing the modified loss function at each iteration
\begin{align*}
&l(\theta;\theta'):=\frac{1}{2}{\mathbb E}_{s,a}[({\mathbb E}_{s',r} [r(s,a) + \gamma J_{\theta'}(s')]-J_{\theta}(s))^2],
\end{align*}
where $\theta$ stands for an online variable and $\theta'$ stands for a target variable. At each iteration step $k$, it sets the target variable to the value of current online variable and performs a stochastic gradient step
\begin{align*}
&\theta_{k+1}=\theta_k-\alpha_k \left. \tilde{\nabla}_{\theta} l(\theta;\theta_k) \right|_{\theta=\theta_k}.
\end{align*}
A full algorithm is described in~\cref{algo:standard-TD-learning}.
\begin{algorithm}[h!]
\caption{Standard TD-Learning}
  \begin{algorithmic}[1]
    \State Initialize $\theta_0$ randomly and set $\theta'_0=\theta_0$.
    \For{iteration $k=0,1,\ldots$}
    	\State Sample $s \sim d(\cdot)$
        \State Sample $a \sim \pi(s,\cdot)$
        \State Sample $s'$ and $r(s,a)$ from SO
        \State Let $g_k = -\phi(s)(r(s,a)+{\gamma}\phi(s')^T\theta'_k-\phi(s)^T \theta_k)$

        \State Update $\theta_{k+1} = \theta_k-\alpha_k g_k$

        \State Update $\theta'_{k + 1}  = \theta_{k+1}$

    \EndFor

  \end{algorithmic}\label{algo:standard-TD-learning}
\end{algorithm}

Inspired by the the recent target-based deep Q-learning algorithms~\citep{mnih2015human}, we consider several alternative updating rules for the target variable that are less aggressive and more general. This then leads to the so-called target-based TD-learning. One of the potential benefits is that by slowing down the update for the target variable,
we can reduce the correlation of the target value, or the variance in the gradient estimation, which would then improve the stability of the algorithm.  To this end, we introduce three variants of target-based TD: averaging TD, double TD, and periodic TD, each of which corresponds to a different strategy of the target update. In the following sections, we discuss these algorithms in details and provide their convergence analysis.

\section{Averaging TD-Learning (A-TD)}
We start by integrating TD-learning  with the Polyak averaging strategy for target variable update. This is motivated by the recent deep Q-learning~\citep{mnih2015human} and DDPG~\citep{lillicrap2015continuous}. It's worth pointing out that such a strategy has been commonly used in the deep Q-learning framework, but the convergence analysis remains absent to our best knowledge. Here we first study this for the TD-learning. The basic idea is to
minimize the modified loss, $l(\theta;\theta')$, with respect to $\theta$ while freezing $\theta'$, and then enforce $\theta' \to \theta$ (target tracking). Roughly speaking, the tracking step, $\theta' \to \theta$, is executed with the update
\begin{align}
&\theta_{k+1}=\theta_{k}-\alpha_k\left. \tilde \nabla_{\theta} l(\theta;\theta'_{k})\right|_{\theta=\theta_{k}}\nonumber\\
& \theta'_{k+1}=\theta'_{k}+\alpha_k \delta (\theta_{k}-\theta'_{k}),\label{eq:target-TD-summary}
\end{align}
where $\delta >0$ is the parameter used to adjust the update speed of the target variable and $\tilde\nabla_{\theta} l(\theta;\theta'_{k})$ is a stochastic estimation of $\nabla_{\theta} l(\theta;\theta'_{k})$. A full algorithm is summarized in~\cref{algo:prototype-algorithm}, which is called \emph{averaging TD} (A-TD).

Compared to the standard TD-learning in~\cref{algo:standard-TD-learning}, the only difference comes from  the target variable update in the last line of~\cref{algo:prototype-algorithm}. In particular, if we set $\alpha_k=1/\delta$ and replace $\theta_k$ with $\theta_{k+1}$ in the second update, then it reduces to the TD-learning.
\begin{algorithm}[h!]
\caption{Averaging TD-Learning (A-TD)}
  \begin{algorithmic}[1]
    \State Initialize $\theta_0$ and $\theta'_0$ randomly.
    \For{iteration $k=0,1,\ldots$}
    	\State Sample $s \sim d(\cdot)$
        \State Sample $a \sim \pi(s,\cdot)$
        \State Sample $s'$ and $r(s,a)$ from SO
        \State Let $g_k =-\phi(s)(r(s,a)+{\gamma}\phi(s')^T\theta'_k-\phi(s)^T \theta_k)$

        \State Update $\theta_{k+1} = \theta_k-\alpha_k g_k$

        \State Update $\theta'_{k + 1}  = \theta'_{k}  + \alpha_k \delta (\theta_{k}  - \theta'_{k})$

    \EndFor
  \end{algorithmic}\label{algo:prototype-algorithm}
\end{algorithm}

Next, we prove its convergence under certain assumptions. The convergence proof is based on the ODE (ordinary differential equation) approach~\citep{bhatnagar2012stochastic}, which is standard technique used in the RL literature~\citep{sutton2009convergent}. In the approach, a stochastic recursive algorithm is converted to the corresponding ODE, and the stability of the ODE is used to prove the convergence. The ODE associated with A-TD is as follows:
\begin{align}
&\dot \theta=-\Phi^T D\Phi\theta+\gamma \Phi^T DP^\pi \Phi\theta'+\Phi^TDR^\pi,\nonumber\\
&\dot\theta'=\delta\theta-\delta\theta'.
\label{eq:ODE-1}
\end{align}

We arrive at the following convergence result,.
\begin{theorem}\label{prop:prototype-convergence}
Assume that with a fixed policy $\pi$, the Markov chain is
ergodic and the step-sizes satisfy
\begin{align}
&\alpha_k>0,\quad \sum_{k=0}^\infty
{\alpha_k}=\infty,\quad \sum_{k=0}^\infty{\alpha_k^2}<\infty.\label{eq:step-size-rule}
\end{align}
Then, $\theta'_k \to \theta^*$ and $\theta_k \to \theta^*$ as $k \to \infty$ with
probability one, where
\begin{align}\label{eq:optimum}
&\theta^*=-(\Phi^T D(\gamma P^\pi -I)\Phi)^{-1} \Phi^T D R^\pi.
\end{align}
\end{theorem}
\begin{remark}
Note that $\theta^*$ in~\eqref{eq:optimum} is not identical to the optimal solution of the original problem in~\eqref{eq:loss-function1}. Instead, it is the solution of the projected Bellman equation defined as
\begin{align*}
&\Phi\theta={\bf F}(\Phi\theta),
\end{align*}
where ${\bf F}$ is the projected Bellman operator defined by
\begin{align*}
&{\bf F}(\Phi\theta):=\Pi(R^\pi+\gamma P^\pi\Phi\theta),
\end{align*}
$\Pi$ is the projection onto the range space of $\Phi$,
denoted by $R(\Phi)$: $\Pi(x):=\argmin_{x'\in R(\Phi)}
\|x-x'\|_D^2$. The projection can be performed by the matrix
multiplication: we write $\Pi(x):=\Pi x$, where $\Pi:=\Phi(\Phi^T
D\Phi)^{-1}\Phi^T D$.
\end{remark}

\cref{prop:prototype-convergence} implies that both the target and online variables of the A-TD converge to $\theta^*$ which solves the projected Bellman equation. The proof of~\cref{prop:prototype-convergence} is provided in~\cref{app:thm1} based on the stochastic approximation approach, where  we apply the Borkar and Meyn theorem~\citep[Appendix~D]{bhatnagar2012stochastic}. Alternatively, the multi-time scale stochastic approximation~\citep[pp.~23]{bhatnagar2012stochastic} can be used with slightly different step-size rules. Due to the introduction of target variable updates, deriving a finite-sample analysis for the modified TD-learning is far from straightforward~\citep{dalal2018finite,bhandari2018finite}. We will leave this for future investigation.

\section{Double TD-Learning (D-TD)}
In this section, we introduce a natural extension of the A-TD, which has a more symmetric form. The algorithm mirrors the double Q-learning~\citep{van2016deep}, but with a notable difference. Here, both the online variable and target variable are updated in the same fashion by switching roles. To enforce $\theta'\to\theta$, we also add a correction term $\delta(\theta-\theta')$ to the gradient update. The algorithm is summarized in~\cref{algo:double-TD-algorithm}, and referred to as the \emph{double TD}-learning (D-TD).
\begin{algorithm}[h!]
\caption{Double TD-Learning (D-TD)}
  \begin{algorithmic}[1]
    \State Initialize $\theta_0$ and $\theta'_0$ randomly.
    \For{iteration $k=0,1,\ldots$}
    	\State Sample $s \sim d(\cdot)$
        \State Sample $a \sim \pi(s,\cdot)$
        \State Sample $s'$ and $r(s,a)$ from SO
        \State Let $g_k = -\phi(s)(r(s,a)+{\gamma}\phi(s')^T\theta'_k-\phi(s)^T \theta_k) - \delta (\theta'_k-\theta_k)$
        \State Let $g'_k = -\phi(s)(r(s,a)+{\gamma}\phi(s')^T\theta_k-\phi(s)^T \theta'_k) - \delta (\theta_k-\theta'_k)$
        \State Update $\theta_{k+1} =\theta_k-\alpha_k g_k$
        \State Update $\theta'_{k+1} =\theta'_k-\alpha_k g'_k$

    \EndFor

  \end{algorithmic}\label{algo:double-TD-algorithm}
\end{algorithm}

We provide the convergence of the D-TD with linear function approximation below.  The proof is similar to the proof of~\cref{prop:prototype-convergence}, and is contained in~\cref{app:thm2}. Noting that asymptotic convergence for double Q-learning has been established in~\cite{hasselt2010double} for tabular case, but no result is yet known when linear function approximation is used.

\begin{theorem}\label{prop:double-TD-convergence}
Assume that with a fixed policy $\pi$, the Markov chain is
ergodic and the step-sizes satisfy~\eqref{eq:step-size-rule}. Then, $\theta_k
\to \theta^*$ and $\theta'_k \to \theta^*$ as $k \to \infty$ with
probability one.
\end{theorem}

If D-TD uses identical initial values for the target and online variables, then the two updates remain identical, i.e., $\theta_k = \theta'_k$ for $k\geq 0$. In this case, D-TD is equivalent to the TD-learning with a variant of the step-size rule. In practice, this problem can also be resolved if we use different samples for each update, and the convergence result will still apply to this variation of D-TD.

Compared to the corresponding form of the double Q-learning~\citep{hasselt2010double}, D-TD has two modifications. First, we introduce an additional term, $\delta (\theta'_k-\theta_k)$ or $\delta (\theta_k-\theta'_k)$, linking the target and online parameter to enforce a smooth update of the target parameter. This covers double Q-learning as a special case by setting $\delta=0$. Moreover, the D-TD updates both target and online parameters in parallel instead of randomly. This approach makes more efficient use of the samples in a slight sacrifice of the computation cost.  The convergence of the randomized version is proved with slight modification of the corresponding proof (see~\cref{app:randomized-D-TD} for details).


\section{Periodic TD-Learning (P-TD)}

In this section, we propose another version of the target-based TD-learning algorithm, which more resembles that used in the deep Q-learning~\citep{mnih2015human}. It corresponds to the periodic update form of the target variable, which differs from  previous sections. Roughly speaking, the target variable is only  periodically updated as follows:
\begin{align*}
&\theta_{k+1}=\theta_k-\alpha_k \left. \tilde \nabla_{\theta} l(\theta;\theta_{k-(k\bmod L)}) \right|_{\theta=\theta_k},
\end{align*}
where $\tilde \nabla_{\theta} l(\theta;\theta_{k-(k\bmod L)})$ is a stochastic estimator of the gradient $\nabla_{\theta} l(\theta;\theta_{k-(k\bmod L)} )$. The standard TD-learning is recovered by setting $L=1$.

Alternatively, one can interpret every $L$ iterations of the update as contributing to minimizing the  modified loss function
$$\min_{\theta} l(\theta;\theta'):=\frac{1}{2}{\mathbb E}_{s,a}[({\mathbb E}_{s',r} [r(s,a) + \gamma J_{\theta'}(s')]-J_{\theta}(s))^2],$$
while freezing the target variable. In other words, the above subproblem is approximately solved at each iteration through $L$ steps of stochastic gradient descent. We formally present the algorithmic idea in a more general way as depicted in~\cref{algo:stochastic-algorithm} and call it the \emph{periodic TD} algorithm (P-TD).

\begin{algorithm}[h!]
\caption{Periodic TD-Learning (P-TD)}
  \begin{algorithmic}[1]
    \State Initialize $\theta_0$ randomly and set $\theta'_0=\theta_0$.
    \State Set positive integers $T$ and the subroutine iteration steps, $L_k$, for $k=0,1,\ldots,T-1$.
    \State Set stepsizes, $\{\beta_t\}_{t=0}^\infty$, for the subproblem.
    \For{iteration $k=0,1,\ldots, T-1$}
    	\State Update
    	\vspace{-2mm}
    \begin{align*}
    \theta_{k+1}= {\tt SGD}(\theta_{k},\theta'_k,L_k)
    \end{align*}
    \vspace{-3mm}
    such that
    \vspace{-2mm}
    $${\mathbb E}[\| \theta_{k+1}-\theta_{k+1}^*\|_2^2] \le \varepsilon_{k + 1},$$
    where $\theta_{k+1}^*:=\argmin_{\theta\in \Theta} l(\theta;\theta'_k)$.

    \State Update $\theta'_{k+1} = \theta_{k+1}$
    \EndFor
    \State {\bf Return} $\theta_{T+1}$
    \item[]

    \Procedure{\tt SGD}{$\theta_k$,$\theta'_k$,$L_k$}
    \item[] \Comment Subroutine: Stochastic gradient decent steps
       \State Initialize $\theta_{k,0}= \theta_k$.
        \For{iteration $t=0,1,\ldots, L_k-1$}
    	\State Sample $s \sim d(\cdot)$
        \State Sample $a \sim \pi(s,\cdot)$
        \State Sample $s'$ and $r(s,a)$ from SO
        \State Let $g_t=-\phi(s)(r(s,a)+{\gamma}\phi(s')^T\theta'_k-\phi(s)^T\theta_{k,t})$
        \State Update $\theta_{k,t+1}=\theta_{k,t}-\beta_t g_t$

    \EndFor
        \State {\bf Return} $\theta_{k,L_k}$
    \EndProcedure
  \end{algorithmic}\label{algo:stochastic-algorithm}
\end{algorithm}

For the P-TD, given a fixed target variable $\theta'_k$, the subroutine, ${\tt SGD} (\theta_k,\theta'_k,L_k)$, runs stochastic gradient descent steps $L_k$ times in order to approximately solve the subproblem $\argmin_{\theta\in {\mathbb R}^n} l(\theta;\theta'_k)$,  for which an unbiased stochastic gradient estimator is obtained by using observations. Upon solving the subproblem after $L_k$ steps, the next target variable is replaced with the next online variable. This makes it similar to the original deep Q-learning~\citep{mnih2015human} as it is periodic if $L_k$ is set to a constant. Moreover, P-TD is also closely related to the TD-learning~\cref{algo:standard-TD-learning}. In particular,
if $L_k=0$ for all $k = 0,1,\ldots,T-1$, then P-TD corresponds to the standard TD.

Based on the standard results in~\citet[Theorem~4.7]{bottou2018optimization}, the ${\tt SGD}$ subroutine converges to the optimal solution, $\theta_{k+1}^*:=\argmin_{\theta\in {\mathbb R}^n} l(\theta;\theta'_k)$. But as we only apply a finite number $L_k$ steps of ${\tt SGD}$, the subroutine will return an approximate solution with a certain error bound $\varepsilon_k$ in expectation, i.e., ${\mathbb E}[\| \theta_{k+1}-\theta_{k+1}^*\|_2^2 |\theta_k] \le \varepsilon_{k + 1}$.

In the following, we establish a finite-time convergence analysis of P-TD. We first present a result in terms of the expected error of the solution and its bounds with high probability.
\begin{theorem}\label{prop:convergence-stochastic-GTD}
Consider~\cref{algo:stochastic-algorithm}. We have
\begin{align*}
{\mathbb E}[\|\Phi \theta_T-\Phi\theta^*\|_D ]
& \le \|\Phi\|_D \sqrt{\max_{s \in {\cal S}} d(s)}\sum_{k=1}^{T-1}{\gamma^{T-k}\sqrt {\varepsilon_k}}+ \gamma^T {\mathbb E}[\| \Phi\theta_0-\Phi\theta^*\|_D ].
\end{align*}
Moreover,
\begin{align*}
{\mathbb P}[\| \Phi\theta_T-\Phi\theta^*\|_D\ge\tau]
&\le \frac{\gamma^T {\mathbb E}[\|\Phi\theta_0-\Phi\theta^*\|_D]}{\tau}+\frac{\|\Phi\|_D\sqrt{\max_{s \in {\cal S}} d(s)}}{\tau }\sum_{k=1}^{T-1}{\gamma^{T-k} \sqrt{\varepsilon_k}}.
\end{align*}
\end{theorem}

The second result implies that P-TD achieves
an $\epsilon$-optimal solution with an arbitrarily high probability by approaching $T \to \infty$ and controlling the error bounds $\varepsilon_k$. In particular, if $\varepsilon_k=\varepsilon$ for all $k \geq 0$, then
\begin{align*}
&{\mathbb P}[\|\Phi\theta_T-\Phi\theta^*\|_D\ge\tau]\le \frac{\|\Phi\|_D \sqrt{\max_{s \in {\cal S}} d(s)}\sqrt\varepsilon}{\tau(1-\gamma)} + \frac{\gamma^T {\mathbb E}[\|\Phi\theta_0-\Phi\theta^*\|_D]}{\tau}.
\end{align*}
One can see that the error is essentially decomposed into two terms, one from the approximation errors induced from SGD procedures and one from the contraction property of solving the subproblems, which can also be viewed as solving the projected Bellman equations. Full details of the proof can be found in~\cref{app:thm3}.

To further analyze the approximation error from the SGD procedure,  existing convergence results in~\citet[Theorem~4.7]{bottou2018optimization} can be applied with slight modifications.

\begin{proposition}\label{prop:convergence-SGD}
Suppose that the SGD method in~\cref{algo:stochastic-algorithm} is run with a stepsize sequence such that, for all $t\geq 0$,
\begin{align*}
\beta_t=\frac{\beta}{\kappa+t+1}
\end{align*}
for some $\beta>1/\lambda_{\min}(\Phi^T D\Phi)$ and $\kappa >0$ such that
\begin{align*}
&\beta_0 =\frac{\beta}{\kappa+1}\le \frac{1}{\sqrt{\lambda_{\max}(\Phi^T D\Phi\Phi^T D\Phi)}(\xi_3+1)},
\end{align*}
Then, for all $0\leq t\leq L_k-1$, the expected optimality gap satisfies
\begin{align}
{\mathbb E}[\|\theta_{k+1}^*-\theta_{k,t}\|_2^2 |\theta_k]\le \frac{2}{\lambda_{\min}(\Phi^T D\Phi)}\frac{\chi_1+\chi_2\|\theta_k -\theta^*\|_2^2}{\kappa+t+1},\label{eq:12}
\end{align}
where
\begin{align*}
\chi_1:=&(\xi_1+\xi_2\|\theta^*\|_2^2)\chi_3 + (\kappa+1)\| R^\pi+P^\pi\Phi\theta^*-\Phi\theta^*\|_D^2,\\
\chi_2:=&\frac{\xi_2\chi_3}{2(\beta\lambda_{\min} (\Phi^TD\Phi)-1)}+(\kappa+1)\lambda_{\max}((P^\pi\Phi-\Phi)^T D(P^\pi\Phi-\Phi)),\\
\chi_3:=&\frac{\beta^2\sqrt{\lambda_{\max}(\Phi^T D\Phi\Phi^T D\Phi)}}{2(\beta\lambda_{\min} (\Phi^TD\Phi)-1)},
\end{align*}
and
\begin{align*}
\xi_1:=&3\sigma^2 \|\Phi\|_2^2 + 2(1+\xi_3)^2 \| \Phi^T DR^\pi \|_2^2,\\
\xi_2:=&3\| \Phi \|_2^4+2(1+\xi_3)^2 \lambda_{\max}(\Phi^T(P^\pi)^T D\Phi\Phi^T DP^\pi \Phi),\\
\xi_3:=&\frac{3\|\Phi\|_2^4}{\lambda_{\min}(\Phi^T D\Phi\Phi^T D\Phi)}.
\end{align*}

\end{proposition}

\cref{prop:convergence-SGD} ensures that the subroutine iterate, $\theta_k$, converges to the solution of the subproblem at the rate of ${\cal O}(1/L_k)$. Combining~\cref{prop:convergence-SGD} with~\cref{prop:convergence-stochastic-GTD}, the overall sample complexity is derived in the following proposition. We defer the proofs to~\cref{app:prop_SGD_convergence} and~\cref{app:prop_sample_complexity}.

\begin{proposition}[Sample Complexity]\label{prop:sample-complexity}
The $\epsilon$-optimal solution, ${\mathbb E}[\|\theta_{T}-\theta^*\|_D] \le \epsilon$, is obtained by~\cref{algo:stochastic-algorithm} with SO calls at most $\rho_1(\rho_2\epsilon^{-2}+4\chi_2)\ln(\epsilon^{-1})$, where
\begin{align*}
&\rho_1:=\frac{2\|\Phi\|_D^2}{\lambda_{\min}(\Phi^T D\Phi)^2(1-\gamma)^2 \ln\gamma^{-1}},\\
&\rho_2:=\chi_1\lambda_{\min}(\Phi^T D\Phi)+\chi_2 {\mathbb E}[\|\Phi\theta_0-\Phi\theta^*\|_D^2],
\end{align*}
and $\chi_1$ and $\chi_2$ are defined in~\cref{prop:convergence-SGD}.
\end{proposition}

As a result, the overall sample complexity of P-TD is bounded by $\mathcal{O}((1/\epsilon^2)\ln(1/\epsilon))$. As mentioned earlier, non-asymptotic analysis for even the standard TD algorithm is only recently developed in a few work~\citep{dalal2018finite,bhandari2018finite,srikant2019}. Our sample complexity result on P-TD, which is a target-based TD algorithm, matches with that developed in~\citet{bhandari2018finite} with similar decaying step-size sequence, up to a log factor. Yet, our analysis is much simpler and builds directly upon existing results on stochastic gradient descent. Moreover, from the computational perspective, although P-TD runs in two loops, it is as the efficient as standard TD.

P-TD also shares some similarity with the least squares temporal difference (LSTD,~\citet{Bradtke1996}) and its stochastic approximation variant (fLSTD-SA,~\citet{prashanth14}). LSTD is a batch algorithm that directly estimates the optimal solution as described in~\eqref{eq:optimum} through samples, which can also be viewed as exactly computing the solution to a least squares subproblem. fLSTD-SA alleviates the computation burden by applying the stochastic gradient descent (the same as TD update) to solve the subproblems. The key difference between fLSTD-SA and P-TD lies in that the objective for P-TD is adjusted by the target variables across cycles. Lastly, P-TD is also closely related to and can be viewed as a special case of the least-squares fitted Q-iteration~\citep{Antos2008}. Both of them solves a similar least squares problems using target values. However, for P-TD, we are able to directly apply the stochastic gradient descent to address the subproblems to near-optimality.

\section{Simulations}\label{section:simulations}
In this section, we provide some preliminary numerical simulation results showing the efficiency of the proposed target-based TD algorithms. We stress that the main goal of this paper is to introduce the family of target-based TD algorithms with linear function approximation and provide theoretical convergence analysis for target TD algorithms, as an intermediate step towards the understanding of target-based Q-learning algorithms. Hence, our numerical experiments simply focus on testing the convergence, sensitivity in terms of the tuning parameters of these target-based algorithms, as well as effects of using target variables as opposed to the standard TD-learning.

\begin{figure*}[ht!]
\centering\subfigure[{Error evolution over $[0,3000]$}]{\includegraphics[width=8cm,height=6cm]{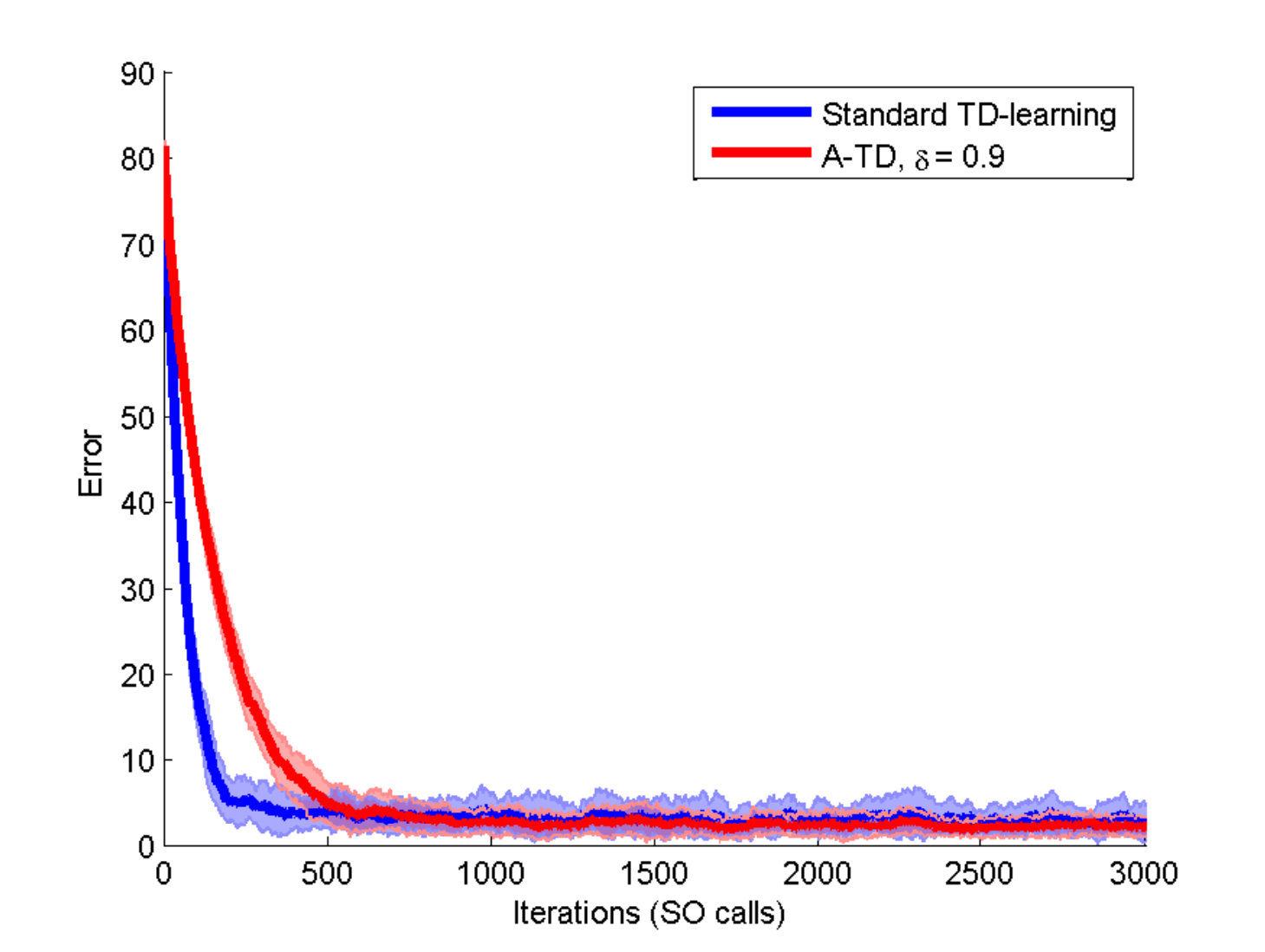}}
\subfigure[{Error evolution over $[2000,3000]$}]{\includegraphics[width=8cm,height=6cm]{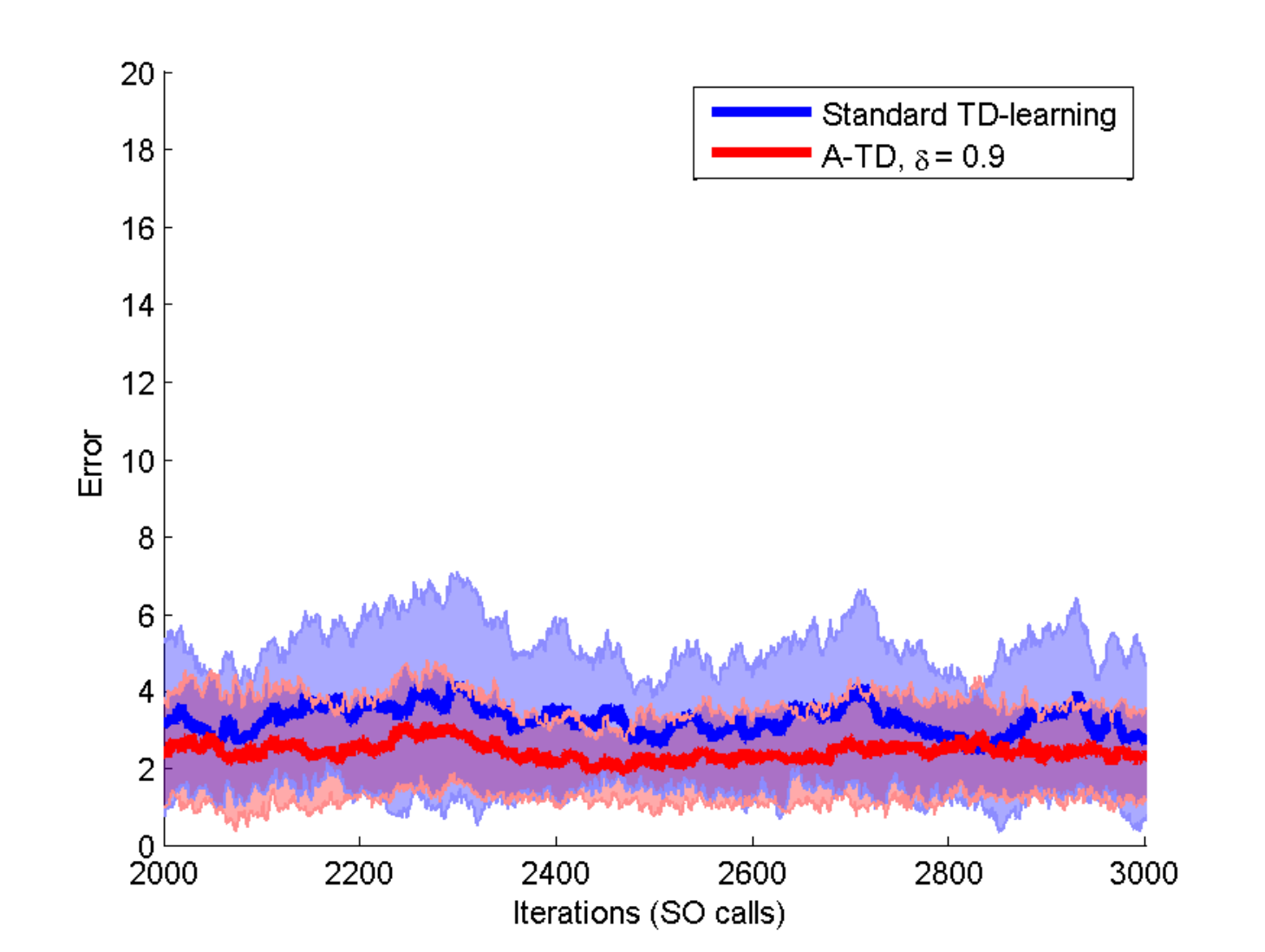}}
\caption{(a) {\color{blue}Blue line}: error evolution of the standard TD-learning with the step-size $\alpha_k=1000/(k+10000)$; {\color{red}Red line}: error evolution of A-TD with the step-size $\alpha_k=1000/(k+10000)$ and $\delta = 0.9$. The shaded areas depict empirical variances obtained with several realizations. (a) Error over the interval $[0,3000]$; (b) Error over the interval $[2000,3000]$.}\label{fig:ex1-figure1}
\end{figure*}
\begin{figure*}[ht!]
\centering\subfigure[{Error evolution over $[0,3000]$}]{\includegraphics[width=8cm,height=6cm]{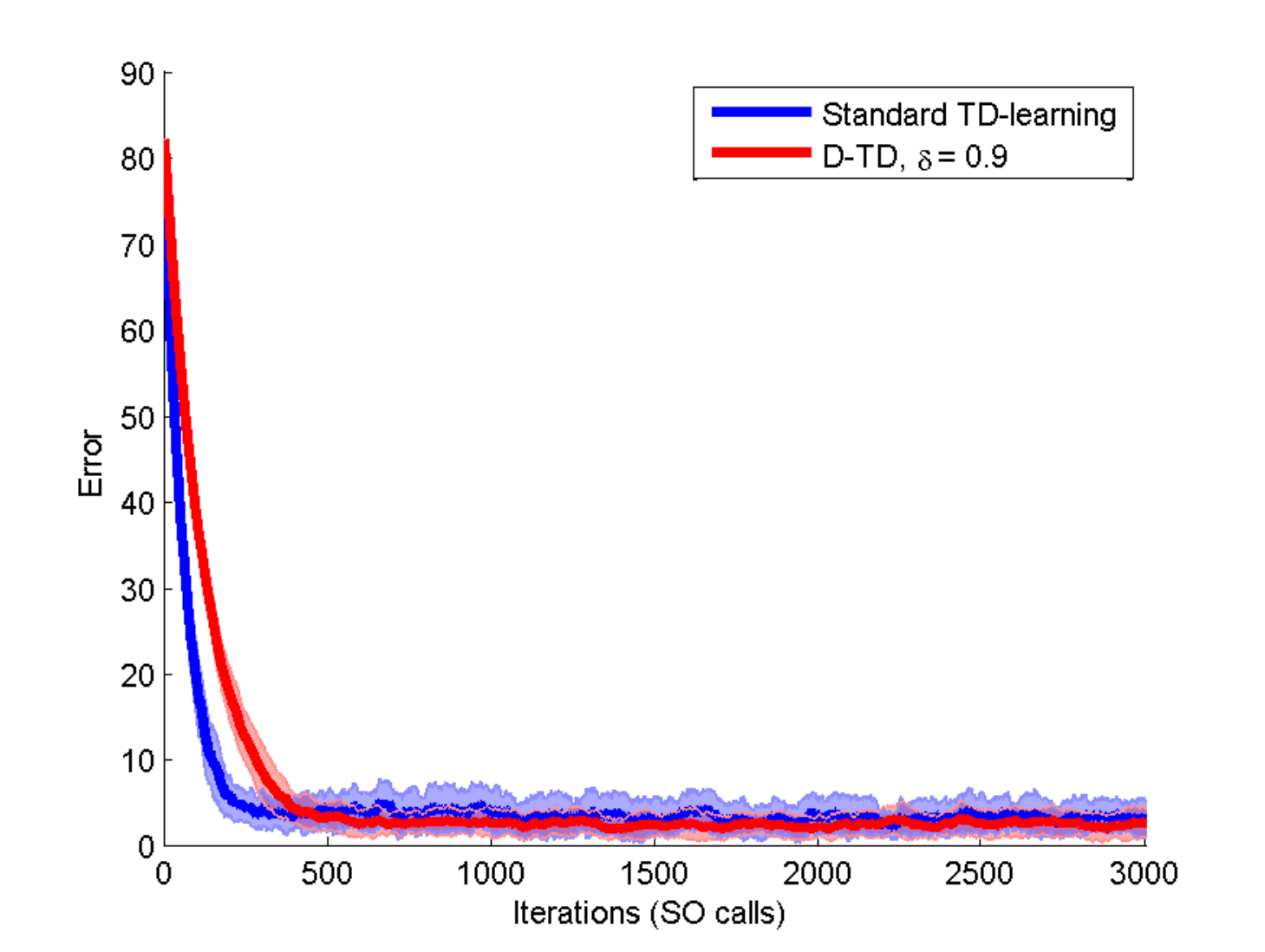}}\subfigure[{Error evolution over $[2000,3000]$}]{\includegraphics[width=8cm,height=6cm]{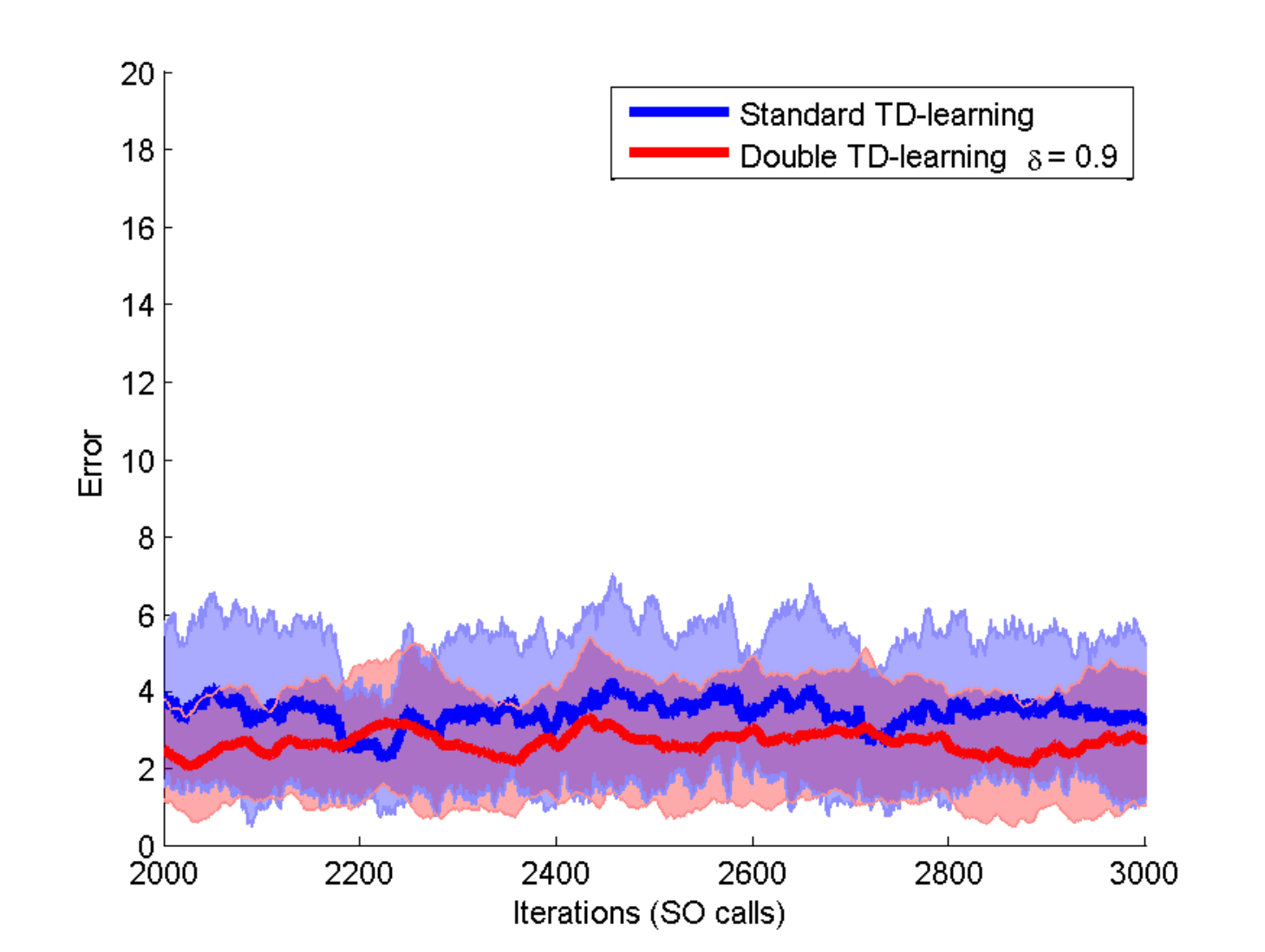}}
\caption{{\color{blue} Blue line}: error evolution of the standard TD-learning with the step-size $\alpha_k=1000/(k+10000)$; {\color{red}Red line}: error evolution of D-TD with the step-size $\alpha_k=1000/(k+10000)$ and $\delta=0.9$. The shaded areas depict empirical variances obtained with several realizations. (a) Error over the interval $[0,3000]$; (b) Error over the interval $[2000,3000]$.}\label{fig:ex1-figure3}
\end{figure*}
\begin{figure*}[ht!]
\centering\subfigure[{Error evolution over $[0,40000]$}]{\includegraphics[width=8cm,height=6cm]{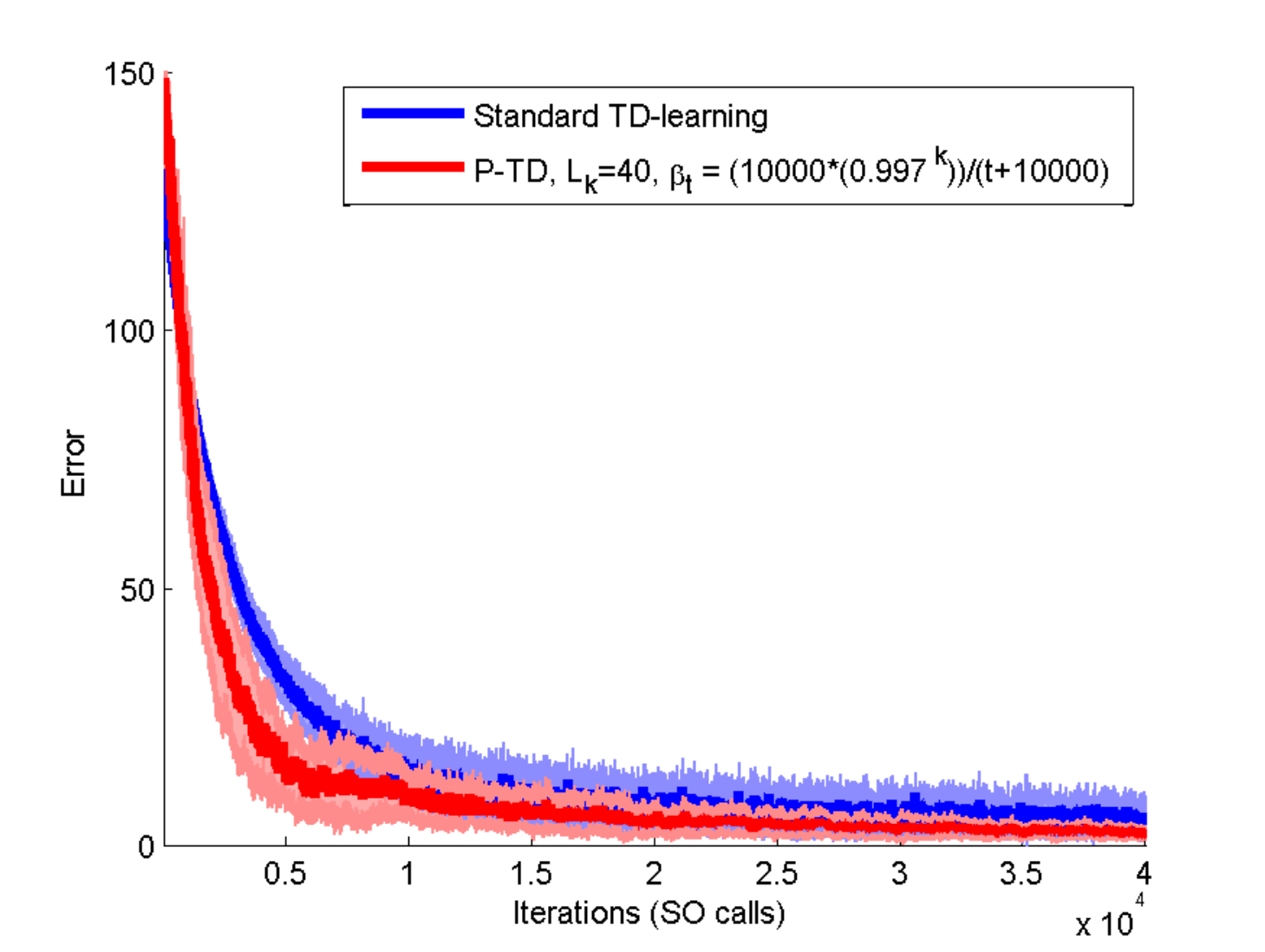}}\subfigure[{Error evolution over $[39000,40000]$}]{\includegraphics[width=8cm,height=6cm]{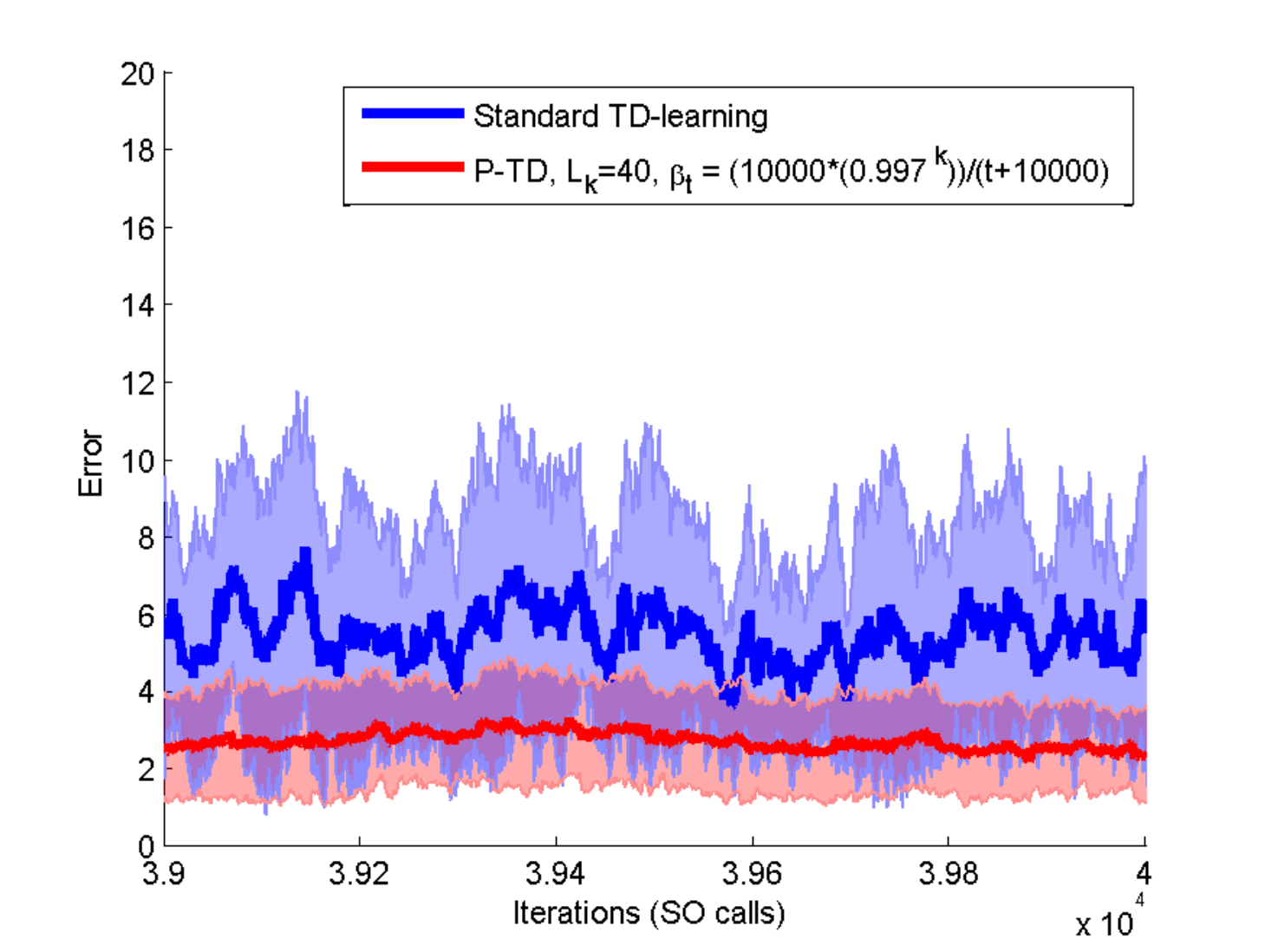}}
\caption{{\color{blue}Blue line}: error evolution of the standard TD-learning with the step-size $\alpha_k=10000/(k + 10000)$. {\color{red}Red line}: error of P-TD with the step-size $\beta_t=(10000\cdot (0.997)^k)/(10000 + t)$ and $L_k=40$. The shaded areas depict empirical variances obtained with several realizations. (a) Error over the interval $[0,30000]$; (b) Error over the interval $[29000,30000]$. }\label{fig:ex2-figure3}
\end{figure*}

\subsection{Convergence of A-TD and D-TD}

In this example, we consider an MDP with $\gamma=0.9$, $|{\cal S}| = 10$,
\begin{align*}
&P^\pi = \begin{bmatrix}
   0.1 & 0.1 & \cdots & 0.1 \\
   0.1 & 0.1 & \ddots & \vdots \\
   \vdots & \ddots & \ddots & 0.1 \\
   0.1 & \cdots & 0.1 & 0.1 \\
\end{bmatrix}\in {\mathbb R}^{10\times 10},
\end{align*}
and $r^\pi(s) \sim U[0,20]$, where $U[0,20]$ denotes the uniform distribution in $[0,20]$ and $r^\pi(s)$ stands for the reward given policy $\pi$ and the current state $s$. The action space and policy are not explicitly defined here. For the linear function approximation, we consider the feature vector with the radial basis function~\citep{geramifard2013tutorial} ($n=2$)
\begin{align*}
&\phi(s) = \begin{bmatrix}
   \frac{\exp(-(s-0)^2)}{2\times 10^2}\\
   \frac{\exp(-(s-10)^2)}{2\times 10^2}\\
\end{bmatrix} \in {\mathbb R}^2.
\end{align*}

Simulation results are given in~\cref{fig:ex1-figure1}, which illustrate error evolution of the standard TD-learning (blue line) with the step-size, $\alpha_k=1000/(k+10000)$ and the proposed A-TD (red line) with the $\alpha_k=1000/(k+10000)$ and $\delta = 0.9$. The design parameters of both approaches are set to demonstrate reasonably the best performance with trial and errors. Additional simulation results in~\cref{app:additional-simulations} provide comparisons for several different parameters. \cref{fig:ex1-figure1}{(b)} provides the results in the same plot over the interval $[2000,3000]$. The results suggest that although A-TD with $\delta = 0.9$ initially shows slower convergence, it eventually converges faster than the standard TD with lower variances after certain iterations. With the same setting, comparative results of D-TD are given in~\cref{fig:ex1-figure3}.

\subsection{Convergence of P-TD}
In this section, we provide empirical comparative analysis of P-TD and the standard TD-learning. The convergence results of both approaches are quite sensitive to the design parameters to be determined, such as the step-size rules and total number of iterations of the subproblem.
We consider the same example as above but with an alternative linear function approximation with the feature vector consisting of the radial basis function
\begin{align*}
&\phi(s) = \begin{bmatrix}
   \frac{\exp(-(s-0)^2)}{2\times 10^2}\\
   \frac{\exp(-(s-10)^2)}{2\times 10^2}\\
   \frac{\exp(-(s-20)^2)}{2\times 10^2}\\
\end{bmatrix} \in {\mathbb R}^3.
\end{align*}

From our own experiences, applying the same step-size rule, $\beta_t$, for every $k \in \{0,1,\ldots,T-1\}$ yields unstable fluctuations of the error in some cases. For details, the reader is referred to~\cref{app:additional-simulations}, which provides comparisons with different design parameters. The results motivate us to apply an adaptive step-size rules for the subproblem of~P-TD so that smaller and smaller step-sizes are applied as the outer-loop steps increases. In particular, we employ the adaptive step-size rule, $\beta_{k,t}=(10000\cdot (0.997)^k)/(10000+t)$ with $L_k=40$ for P-TD, and the corresponding simulation results are given in~\cref{fig:ex2-figure3}, where P-TD outperforms the standard TD with the step-size, $\alpha_k=10000/(k+10000)$, best tuned for comparison. \cref{fig:ex2-figure3}{(b)} provides the results in~\cref{fig:ex2-figure3} in the interval $[29000,30000]$, which clearly demonstrates that the error of P-TD is smaller with lower variances.

\section{Conclusion}

In this paper, we propose a new family of target-based TD-learning algorithms, including the {\em averaging TD, double TD}, and {\em periodic TD}, and provide theoretical analysis on their convergences. The proposed TD algorithms are largely inspired by the recent success of deep Q-learning using target networks and mirror several of the practical strategies used for updating target network in the literature. Simulation results show that integrating target variables into TD-learning can also help stabilize the convergence by reducing variance of and correlations with the target. Our convergence analysis provides some theoretical understanding of target-based TD algorithms. We hope this would also shed some light on the theoretical analysis for target-based Q-learning algorithms and non-linear RL frameworks.


Possible future topics include (1) developing finite-time convergence analysis for A-TD and D-TD; (2) extending the analysis of the target-based TD-learning to the Q-learning case w/o function approximation; and (3) generalizing the target-based framework to other variations of TD-learning and Q-learning algorithms.

\bibliography{reference}
\bibliographystyle{icml2019}

\clearpage
\newpage

\appendices
\onecolumn\

\begin{appendix}

\thispagestyle{plain}
\begin{center}
{\huge Appendix}
\end{center}

\section{Proof of~\cref{prop:prototype-convergence}}\label{app:thm1}
The proof is based on the analysis of the general stochastic
recursion
\begin{align*}
&\theta_{k+1}=\theta_k+\alpha_k (h(\theta_k)+\varepsilon_{k+1}).
\end{align*}
where $h$ is a mapping $h:{\mathbb R}^n \to {\mathbb R}^n$. If only the asymptotic convergence is our concern, the ODE (ordinary differential equation) approach~\citep{bhatnagar2012stochastic} is a convenient tool. Before starting the main proof, we review essential knowledge of the linear system theory~\citep{chen1995linear}.
\begin{definition}[{\citet[Definition~5.1]{chen1995linear}}]
The ODE, $\dot x(t)=Ax(t)$, $t\geq 0$, where $A \in {\mathbb R}^{n\times n}$ and $x(t) \in {\mathbb R}^n$, is asymptotically stable if for every finite initial state $x(0) = x_0$, $x(t)\to 0$ as $t \to \infty$.
\end{definition}
\begin{definition}[Hurwitz matrix]\label{def:Hurwitz-matrix}
A complex square matrix $A \in {\mathbb C}^{n\times n}$ is Hurwitz if all eigenvalues of $A$ have strictly negative real
parts.
\end{definition}
\begin{lemma}[{\citet[Theorem~5.4]{chen1995linear}}]\label{lemma:stability}
The ODE, $\dot x(t)=A x(t)$, $t\geq 0$, is asymptotically stable if and only if $A$ is Hurwitz.
\end{lemma}
\begin{lemma}[Lyapunov theorem {\citep[Theorem~5.5]{chen1995linear}}]\label{lemma:Lyapunov-theorem}
A complex square matrix $A \in {\mathbb C}^{n\times n}$ is Hurwitz if and only if there exists a positive definite matrix $M = M^H \succ 0$ such that $A^H M + MA \prec 0$, where $A^H$ is the complex conjugate transpose of $A$.
\end{lemma}
\begin{lemma}[Schur complement {\citep[pp.~651]{boyd2004}}]\label{lemma:Schur-complement}
For any complex block matrix $\begin{bmatrix}
   A & B  \\
   B^T & C  \\
\end{bmatrix}$, we have
\begin{align*}
&\begin{bmatrix}
   A & B  \\
   B^T & C  \\
\end{bmatrix} \succ 0 \Leftrightarrow A \succ 0,C - B^T A^{-1} B.
\end{align*}
\end{lemma}

Convergence of many RL algorithms rely on the ODE approaches~\citep{bhatnagar2012stochastic}. One of the most popular approach is based on the Borkar and Meyn theorem~\citep[Appendix~D]{bhatnagar2012stochastic}. Basic technical assumptions are given below.
\begin{assumption}\label{assumption:1}
$\,$\begin{enumerate}
\item The mapping $h:{\mathbb R}^n  \to {\mathbb R}^n$ is
globally Lipschitz continuous and there exists a function
$h_\infty:{\mathbb R}^n\to {\mathbb R}^n$ such that
\begin{align*}
&\lim_{c\to \infty}\frac{h(c\theta)}{c}=
h_\infty(\theta),\quad \forall \theta \in {\mathbb R}^n.
\end{align*}

\item The origin in ${\mathbb R}^n$ is an asymptotically stable
equilibrium for the ODE $\dot \theta(t) = h_\infty (\theta(t))$.

\item There exists a unique globally asymptotically stable equilibrium
$\theta^e\in {\mathbb R}^n$ for the ODE $\dot\theta(t) =
h(\theta(t))$, i.e., $\theta(t) \to \theta^e$ as $t \to \infty$.

\item The sequence $\{\varepsilon_k,{\cal G}_k,k\ge 1\} $
with ${\cal G}_k=\sigma(\theta_i,\varepsilon_i,i\le k)$
is a Martingale difference sequence. In addition, there exists a
constant $C_0<\infty $ such that for any initial $\theta_0\in
{\mathbb R}^n$, we have ${\mathbb E}[\|\varepsilon_{k+1}
\|^2 |{\cal G}_k]\le C_0(1+\|\theta_k\|^2),\forall k \ge 0$.

\item The step-sizes satisfy~\eqref{eq:step-size-rule}.
\end{enumerate}
\end{assumption}
\begin{lemma}[Borkar and Meyn theorem]\label{lemma:Borkar}
Suppose that~\cref{assumption:1} holds. For any initial $\theta_0\in
{\mathbb R}^n$, $\sup_{k\ge 0} \|\theta_k\|<\infty$
with probability one. In addition, $\theta_k\to\theta^e$ as
$k\to\infty$ with probability one.
\end{lemma}

Based on the technical results, we are in position to prove~\cref{prop:prototype-convergence}.

{\bf Proof of~\cref{prop:prototype-convergence}}:
The ODE~\eqref{eq:ODE-1} can be expressed as the linear system
with an affine term
\begin{align*}
&\dot{\bar\theta} = A\bar\theta+b=:h\left(\begin{bmatrix}
   \theta\\
   \theta'\\
\end{bmatrix} \right),
\end{align*}
where
\begin{align*}
&A:=\begin{bmatrix}
   -\Phi^T D\Phi & \gamma\Phi^T D P^\pi \Phi\\
   \delta I & -\delta I\\
\end{bmatrix},\quad b:= \begin{bmatrix}
   \Phi^TD R^\pi\\
   0\\
\end{bmatrix},\quad \bar\theta:=\begin{bmatrix}
\theta\\
\theta'\\
\end{bmatrix}.
\end{align*}
Therefore, the mapping $h:{\mathbb R}^n\to {\mathbb R}^n$, defined by $h(\bar\theta)=A\bar\theta + b$, is
globally Lipschitz continuous. Moreover, we have
\begin{align*}
&h_\infty(\bar\theta):=\lim_{t\to \infty} h(t\bar\theta)/t=A\bar\theta.
\end{align*}

Therefore, the first condition in~\cref{assumption:1} holds. To meet the second condition of~\cref{assumption:1}, by~\cref{lemma:stability}, it suffices to prove that $A$ is Hurwitz. The reason is explained below. Suppose that $A$ is Hurwitz. If $A$ is Hurwitz, it is invertible, and there exists a unique equilibrium $\bar\theta^e\in {\mathbb R}^n$ for the ODE $\dot{\bar\theta}= A{\bar\theta}+b$ such that $0=A\bar\theta^e +b$, i.e., $\bar\theta^e=-A^{-1}b$. Due to the constant term $b$, it is not clear if such equilibrium point, $\bar\theta^e$, is globally asymptotically stable. From~\citep[pp.~143]{antsaklis2007linear}, by letting $x=\bar\theta -\bar\theta^e$, the ODE can be transformed to $\dot x = Ax$, where the origin is the globally asymptotically stable equilibrium point since $A$ is Hurwitz. Therefore, $\bar\theta^e$ is globally asymptotically stable equilibrium point of $\dot{\bar\theta}=A{\bar\theta}+b$, and the third condition of~\cref{assumption:1} is satisfied. Therefore, it remains to prove that $A$ is Hurwitz. We first provide a simple analysis and prove that there exists a $\delta^*>0$ such that for all $\delta\geq \delta^*$, $A$ is Hurwitz. To this end, we use the property of the similarity transformation~\citep[pp.~88]{antsaklis2007linear}, i.e., $A$ is Hurwitz if and only if $BAB^{-1}$ is Hurwitz for any invertible matrix $B$. Letting $B = \begin{bmatrix}
I & 0\\
-I & I\\
\end{bmatrix}$, one gets
\begin{align*}
&BAB^{-1}=\begin{bmatrix}
   I & 0  \\
   -I & I  \\
\end{bmatrix}\begin{bmatrix}
   -\Phi^T D\Phi & \gamma\Phi^T DP^\pi \Phi \\
   \delta I & -\delta I\\
\end{bmatrix}\begin{bmatrix}
   I & 0  \\
   I & I  \\
\end{bmatrix} = \begin{bmatrix}
   -\Phi^T D\Phi +\gamma\Phi^T DP^\pi\Phi & \gamma\Phi^T DP^\pi\Phi\\
   \Phi^T D\Phi -\gamma\Phi^T DP^\pi\Phi & -\gamma\Phi^T DP^\pi\Phi -\delta I\\
\end{bmatrix}
\end{align*}

To prove that $BAB^{-1}$ is Hurwitz, we use~\cref{lemma:Lyapunov-theorem} with $M=I$ and check the sufficient condition
\begin{align}
&BAB^{-1}+BA^T B^{-1}\nonumber\\
=& \begin{bmatrix}
   \Phi^T D(-I+\gamma P^\pi)\Phi & \gamma\Phi^T DP^\pi\Phi\\
   -\Phi^T D(-I+\gamma P^\pi)\Phi & -\delta I - \gamma\Phi^T DP^\pi\Phi \\
\end{bmatrix} + \begin{bmatrix}
   \Phi^T D (-I+\gamma P^\pi)\Phi & \gamma\Phi^T DP^\pi\Phi\\
   -\Phi^T D(-I+\gamma P^\pi)\Phi & -\delta I - \gamma \Phi^T DP^\pi\Phi\\
\end{bmatrix}^T\nonumber\\
=& \begin{bmatrix}
   \Phi^T D(-I+\gamma P^\pi)\Phi +\Phi^T(-I+\gamma P^\pi)^T D\Phi & -\Phi^T (-I+\gamma P^\pi)^T D\Phi+\gamma\Phi^T DP^\pi\Phi\\
     -\Phi^T D(-I+\gamma P^\pi)\Phi +\gamma\Phi^T (P^\pi)^T D\Phi & -2\delta I -\gamma\Phi^T DP^\pi \Phi -\gamma\Phi^T (P^\pi)^T D\Phi \\
\end{bmatrix}\nonumber\\
\prec& 0.\label{eq:15}
\end{align}

To check the above matrix inequality, note that $\Phi^T D(\gamma P^\pi-I)\Phi$ is negative definite~\citep[Lemma~6.6, pp.~300]{bertsekas1996neuro}. By using the Schur complement~\cref{lemma:Schur-complement},~\eqref{eq:15} holds if and only if
\begin{align}
0 \prec& 2\delta I + \gamma \Phi ^T DP\Phi  + \gamma \Phi ^T P^T D\Phi\nonumber\\
&- \{  - \Phi ^T ( - I + \gamma P^\pi  )^T D\Phi  + \gamma \Phi ^T DP^\pi  \Phi \} ^T ( - \Phi ^T D( - I + \gamma P)\Phi  - \Phi ^T ( - I + \gamma P)^T \Phi )^{ - 1}\nonumber\\
&\times \{  - \Phi ^T ( - I + \gamma P^\pi  )^T D\Phi  + \gamma \Phi ^T DP^\pi  \Phi \}\label{eq:18}
\end{align}

The above inequality holds for a sufficiently large $\delta$, i.e., there exists $\delta^*>0$ such that the above inequality holds for all $\delta > \delta^*$. Therefore, $BAB^{-1}$ and $A$ are Hurwitz for all $\delta > \delta^*$. A natural question is whether or not $\delta^*=0$. We prove that this is indeed the case. The proof requires rather more involved analysis.

{\bf Claim:} $A$ is Hurwitz for all $\delta >0$.

{\bf Proof:} We investigate the equation
\begin{align*}
&\begin{bmatrix}
   -\Phi^T D\Phi & \gamma \Phi^T DP^\pi \Phi\\
   \delta I & -\delta I \\
\end{bmatrix} \begin{bmatrix}
   x  \\
   y  \\
\end{bmatrix} = \lambda \begin{bmatrix}
   x  \\
   y  \\
\end{bmatrix},
\end{align*}
where $\begin{bmatrix} x \\ y \\
\end{bmatrix}\in {\mathbb C}^{2n}$ is an eigenvector and $\lambda\in {\mathbb C}$ is an eigenvalue of $A$. Equivalently, it is written by
\begin{align}
\lambda x =& -\Phi^T D\Phi x+\gamma\Phi^T DP^\pi\Phi y,\label{eq:19}\\
\lambda y =& \delta (x-y).\label{eq:20}
\end{align}

Solving~\eqref{eq:20} leads to $y=\frac{\delta }{\delta + \lambda}x$, and plugging this expression into $y$ in~\eqref{eq:19} yields
\begin{align}
&\left(-\Phi^T D\Phi+\gamma\frac{\delta}{\delta+\lambda}\Phi^T DP^\pi\Phi\right)x = \lambda x.\label{eq:21}
\end{align}

For any $\delta >0$, the complex number in the above equation
\begin{align}
s:=\frac{\delta}{\delta+\lambda} = \frac{\delta(\lambda^*+\delta)}{|\lambda+\delta |^2}\in {\mathbb C},
\end{align}
where $\lambda^*$ is the complex conjugate of $\lambda \in {\mathbb C}$ and $|\cdot|$ is the absolute value of a complex number $(\cdot)$, has the absolute value less than or equal to $1$, i.e., $|s|=\frac{\delta}{|\lambda +\delta|} < 1$. Now, we prove that the complex matrix, $\Phi^T D(-I+s\gamma P^\pi)\Phi$, is Hurwitz for any $s \in {\mathbb C}$ such that $|s|\leq 1$. For any real vector $v \in {\mathbb R}^{|{\cal S}|}$, we have
\begin{align*}
v^T (\gamma sDP^\pi+\gamma s^* (P^\pi)^T D)v =&\gamma (s+s^*)v^T D^{1/2} D^{1/2} P^\pi v\\
\le& \gamma (s+s^*) \|D^{1/2} v \|_2 \|D^{1/2} P^\pi v\|_2\\
=& \gamma (s+s^*) \|v\|_D \|P^\pi v\|_D\\
\le& \gamma (s+s^*) \|v\|_D \|v\|_D\\
=& \gamma (s+s^*) \| v \|_D^2\\
=& \gamma (s+s^*)v^T Dv\\
\le& \gamma 2v^T Dv,
\end{align*}
where the first inequality is due to the Cauchy-Schwarz inequality, the second inequality is due to~\citet[Lemma~1]{tsitsiklis1997analysis}, and the final inequality follows from the fact that $|s|\leq 1$ implies $-2 \leq s+s^* \leq 2$. The last result ensures $v^T (s\gamma DP^\pi +s^* \gamma (P^\pi)^T D)v \preceq \gamma 2v^T Dv$ for any $v \in {\mathbb R}^{|{\cal S}|}$, and equivalently,
\begin{align*}
D(-I+s\gamma P^\pi)+(-I+s^*\gamma (P^\pi)^T)D \preceq 2(\gamma-1)D.
\end{align*}

Multiplying both sides of the above inequality by $\Phi$ from the right and its transpose from the left, one gets
\begin{align*}
&\Phi^T D(-I+ s\gamma P^\pi)\Phi +\Phi^T (-I+s^* \gamma (P^\pi)^T )D\Phi \preceq 2(-1+\gamma )\Phi^T D\Phi \prec 0.
\end{align*}

By~\cref{lemma:Lyapunov-theorem} with $M=I$, we conclude that the complex matrix, $\Phi^T D(-I+ s\gamma P^\pi)\Phi$, is Hurwitz for any $s\in {\mathbb C}$ such that $|s|\leq 1$. Based on this observation, we return to~\eqref{eq:21} and conclude that $-\Phi^T D\Phi+\gamma\frac{\delta}{\delta+\lambda}\Phi^T DP^\pi\Phi$ is Hurwitz for any $\lambda \in {\mathbb C}$. By the definition of a Hurwitz matrix in~\cref{def:Hurwitz-matrix} and the eigenvalue, we conclude that the real part of $\lambda$ should be always strictly negative. Therefore, $A$ is Hurwitz for any $\delta >0$. This completes the proof. $\quad \blacksquare$.

Next, we prove the remaining parts. Since $\varepsilon_{k+1}$ can be expressed as an affine map of $\bar\theta_k=[\bar\theta_k,\theta'_k]^T$, it can be easily proved that the fourth condition of~\cref{assumption:1} is satisfied. In particular, if we define $m_k :=\sum_{i=0}^k {\varepsilon_i}$, then $m_k$ is Martingale, and $\varepsilon_k$ is a Martingale difference sequence. Therefore, the fourth condition is met.

Finally, by~\cref{lemma:Borkar}, $\bar\theta_k$ converges to $\bar\theta^e$ such that
\begin{align*}
&h(\bar\theta)= \begin{bmatrix}
   -\Phi^T D\Phi & \gamma\Phi^T DP^\pi \Phi\\
   \delta I & -\delta I \\
\end{bmatrix} \begin{bmatrix}
   \theta\\
   \theta'\\
\end{bmatrix}+\begin{bmatrix}
   \Phi^T DR^\pi\\
   0\\
\end{bmatrix}= 0.
\end{align*}

By the block matrix inversion, solving the equation
leads to the desired conclusion, i.e., $\bar\theta^e=\begin{bmatrix} \theta^*\\ \theta^* \end{bmatrix}$.

\vskip 0.2in

\section{Proof of~\cref{prop:double-TD-convergence}}\label{app:thm2}

The ODE corresponding to~\cref{algo:double-TD-algorithm} can be expressed as the linear system with an affine term
\begin{align*}
&\dot{\bar\theta} = A\bar \theta + b=:h\left( \begin{bmatrix}
   \theta\\
   \theta'\\
\end{bmatrix} \right),
\end{align*}
where
\begin{align*}
&A:=\begin{bmatrix}
   -\Phi^T D\Phi- \delta I & \alpha \Phi^T DP^\pi \Phi +\delta I\\
   {\alpha \Phi^T DP^\pi  \Phi  + \delta I} & -\Phi^T D\Phi -\delta I\\
\end{bmatrix},\quad b:= \begin{bmatrix}
   \Phi^TD R^\pi\\
   \Phi^TD R^\pi\\
\end{bmatrix},\quad \bar \theta:=\begin{bmatrix}
\theta\\
\theta'\\
\end{bmatrix}.
\end{align*}

The proof follows the same lines as the proof of~\cref{prop:prototype-convergence}. Therefore, we only prove that $A$ is Hurwitz here. In particular, $A$ can be represented by $A=B+C^T BC$, where
\begin{align*}
&B =\begin{bmatrix}
   -\Phi^T D\Phi & \Phi^T DP^\pi \Phi\\
   \delta I & -\delta I \\
\end{bmatrix},\quad C = \begin{bmatrix}
   0 & I \\
   I & 0 \\
\end{bmatrix}.
\end{align*}

From the proof of~\cref{prop:prototype-convergence}, $B$ is Hurwitz, and admits the Lyapunov matrix $M=I$ such that $B^TM+MB \prec 0$. Thus, $\begin{bmatrix} B & 0 \\ 0 & B\\ \end{bmatrix}$ is Hurwitz as well, and
\begin{align*}
&\begin{bmatrix}
   B & 0  \\
   0 & B  \\
\end{bmatrix}^T+\begin{bmatrix}
   B & 0  \\
   0 & B  \\
\end{bmatrix} \prec 0.
\end{align*}

Pre- and post-multiplying the left-hand side of the about inequality by the full rank matrix $\begin{bmatrix} I & C^T\\
\end{bmatrix}$ and its transpose, respectively, yields
\begin{align*}
\begin{bmatrix}
 I\\
 C\\
\end{bmatrix}^T \begin{bmatrix}
 B & 0\\
 0 & B\\
\end{bmatrix}^T \begin{bmatrix}
   I  \\
   C  \\
\end{bmatrix}+\begin{bmatrix}
 I\\
 C\\
\end{bmatrix}^T \begin{bmatrix}
 B & 0\\
 0 & B\\
\end{bmatrix} \begin{bmatrix}
   I  \\
   C  \\
\end{bmatrix}=B + CBC + B^T  + CB^T C = A^T + A \prec 0.
\end{align*}

By~\cref{lemma:Lyapunov-theorem} with $M=I$, this implies that $A$ is Hurwitz. This completes the proof.

\vskip 0.2in

\section{Randomized version of D-TD}\label{app:randomized-D-TD}

We consider a randomized version of D-TD in~\cref{algo:double-TD-algorithm-random}, which updates either the target or online parameters randomly.
\begin{algorithm}[h!]
\caption{Double TD-Learning (D-TD) with Random Update}
  \begin{algorithmic}[1]
    \State Initialize $\theta_0$ and $\theta'_0$ randomly.
    \For{iteration $k=0,1,\ldots$}
    	\State Sample $s \sim d(\cdot)$
        \State Sample $a \sim \pi(s,\cdot)$
        \State Sample $s'$ and $r(s,a)$ from SO
        \State Choose UPDATE(A) with probability $\nu \in (0,1)$ and UPDATE(B) with probability $1-\nu$
        \If{UPDATE(A)}
            \State Let $g_k = -\phi(s)(r(s,a)+{\gamma}\phi(s')^T\theta'_k-\phi(s)^T \theta_k) + \delta (\theta'_k-\theta_k)$
            \State Update $\theta_{k+1} =\theta_k-\alpha_k g_k$
        \ElsIf{UPDATE(B)}
            \State Let $g'_k = -\phi(s)(r(s,a)+{\gamma}\phi(s')^T\theta_k-\phi(s)^T \theta'_k) + \delta (\theta_k-\theta'_k)$
            \State Update $\theta'_{k+1} =\theta'_k-\alpha_k g'_k$
        \EndIf

    \EndFor

  \end{algorithmic}\label{algo:double-TD-algorithm-random}
\end{algorithm}

We have the convergence result similar to~\cref{prop:double-TD-convergence}.
\begin{theorem}\label{prop:double-TD-convergence=random}
Consider~\cref{algo:double-TD-algorithm-random} and assume that with a fixed policy $\pi$, the Markov chain is
ergodic and the step-sizes satisfy~\eqref{eq:step-size-rule}. Then, $\theta_k
\to \theta^*$ and $\theta'_k \to \theta^*$ as $k \to \infty$ with
probability one.
\end{theorem}
{\bf Proof:} The proof is a slight modification of the proof of~\cref{prop:double-TD-convergence}. The ODE corresponding to~\cref{algo:double-TD-algorithm-random} can be expressed as the linear system with an affine term
\begin{align*}
&\dot{\bar\theta} =\Lambda A\bar \theta + b=:h\left( \begin{bmatrix}
   \theta\\
   \theta'\\
\end{bmatrix} \right),
\end{align*}
where $A$ is defined in~\cref{app:thm2} and $\Lambda = \begin{bmatrix} \nu I & 0 \\ 0 & (1-\nu) I\\ \end{bmatrix}$. The remaining part is to prove that $\Lambda A$ is Hurwitz. From the proof of~\cref{prop:double-TD-convergence}, we know $A^T + A \prec 0$, which is equivalent to $(A^T \Lambda) \Lambda^{-1} + \Lambda^{-1} (\Lambda A) \prec 0$. By~\cref{lemma:Lyapunov-theorem} with $M=\Lambda^{-1}$, this implies that $\Lambda A$ is Hurwitz. This completes the proof. $\quad\blacksquare$

\vskip 0.2in

\section{Proof of~\cref{prop:convergence-stochastic-GTD}}\label{app:thm3}

Before presenting the proof, we first introduce a deterministic version of P-TD summarized in~\cref{algo:deterministic-algorithm} in order to make smooth steps forward. For a fixed $\theta'_k$ (target variable), the subroutine, ${\tt GradientDecent}$, runs gradient descent steps $L_k$ times in order to approximately solve the subproblem, $\argmin_{\theta\in {\mathbb R}^n} l(\theta;\theta'_k)$. By the standard results in~\citet[Theorem~10.3]{bubeck2015convex}, the gradient descent iterations converge to the optimal solution $\theta_{k+1}^*:=\argmin_{\theta\in {\mathbb R}^n} l(\theta;\theta'_k)$ linearly, the finite iterates reache an approximate solution within a certain error bound $\varepsilon_{k}$. Upon solving the subproblem, the next target variable is replaced with the next online variable.
\begin{algorithm}[h!]
\caption{Deterministic Periodic TD-Learning}
  \begin{algorithmic}[1]
    \State Initialize $\theta_0$ randomly and set $\theta'_0=\theta_0$
    \State Set positive integers $T$ and $L_k$ for $k=0,1,\ldots,T-1$
    \State Set stepsizes, $\{\beta_t\}_{t=0}^\infty$, for the subproblem
    \For{iteration $k=0,1,\ldots, T-1$}
    	\State Update
    \begin{align*}
    &\theta_{k+1} = {\tt GradientDecent}(\theta_{k},\theta'_{k},L_k)
    \end{align*}
    such that
    \begin{align*}
    &\|\theta_{k+1}-\theta_{k+1}^* \|_2^2\le\varepsilon_{k+1},
    \end{align*}
    where $\varepsilon_k>0$ is an error bound and $\theta_{k+1}^*:=\argmin_{\theta\in {\mathbb R}^n} l(\theta;\theta'_k)$.
    \State Update $\theta'_{k+1} = \theta_{k+1}$
    \EndFor
    \State {\bf Return} $\theta_{T+1}$

    \item[]

    \Procedure{GradientDecent}{$\theta_k$,$\theta'_k$,$L_k$}
        \item[] \Comment Subroutine: Gradient decent steps
        \State Set $\theta_{k,0} = \theta_k$
        \For{iteration $t=0,1,\ldots,L_k-1$}
            \State Update
            \begin{align*}
            &\theta_{k,t+1}=\theta_{k,t}-\beta_t \left. \nabla_{\theta} l(\theta;\theta'_k) \right|_{\theta=\theta_{k,t}}.
            \end{align*}
        \EndFor
        \State {\bf Return} $\theta_{k,L_k}$
    \EndProcedure
  \end{algorithmic}\label{algo:deterministic-algorithm}
\end{algorithm}

The overall convergence relies on the fact that approximately solving the subproblem can be interpreted as approximately solving a projected Bellman equation defined below.
\begin{definition}[Projected Bellman equation]
The projected Bellman equation is defined as
\begin{align*}
&\Phi\theta= {\bf F}(\Phi\theta),
\end{align*}
where ${\bf F}$ is the projected Bellman operator defined by
\begin{align*}
&{\bf F}(\Phi\theta):=\Pi(R^\pi+\gamma P^\pi\Phi\theta),
\end{align*}
$\Pi$ is the projection onto the range space of $\Phi$,
denoted by $R(\Phi)$: $\Pi(x):=\argmin_{x'\in R(\Phi)}
\|x-x'\|_D^2$. The projection can be performed by the matrix
multiplication: we write $\Pi(x):=\Pi x$, where $\Pi:=\Phi(\Phi^T
D\Phi)^{-1}\Phi^T D$.
\end{definition}

By direct calculations, we can conclude that the solution of the projected Bellman equation is not identical to the solution of the value function evaluation problem in~\eqref{eq:loss-function1}, while it only approximates the solution of~\eqref{eq:loss-function1}. The solution of the projected Bellman equation is denoted by $\theta^*$, i.e.,
\begin{align*}
&\Phi \theta^*= {\bf F}(\Phi\theta^*).
\end{align*}

Therefore,~\cref{algo:deterministic-algorithm} executes an approximate dynamic programming procedure. Based on these observations, the convergence of~\cref{algo:deterministic-algorithm} is given below.
\begin{proposition}\label{prop:convergence-deterministic-GTD}
Consider~\cref{algo:deterministic-algorithm}. We have
\begin{align*}
&\| \Phi\theta_T- \Phi\theta^*\|_D\le \sqrt{\max_{s\in {\cal S}}d(s)} \|\Phi\|_D \sum_{k=1}^{T} {\gamma^{T-k} \sqrt{\varepsilon_k}}+ \gamma^T\|\Phi\theta_0-\Phi\theta^*\|_D.
\end{align*}
\end{proposition}

To prove~\cref{prop:convergence-deterministic-GTD}, we first summarize some essential technical lemmas. The first lemma states that the operator ${\bf F}$ is a contraction.
\begin{lemma}\label{lemma:contraction}
The operator ${\bf F}$ is a $\gamma$-contraction with respect to $\|\cdot\|_D$, i.e.,
\begin{align*}
&\|{\bf F}(\Phi x)-{\bf F}(\Phi y)\|_D \le \gamma \|\Phi x -\Phi y\|_D.
\end{align*}
\end{lemma}
{\bf Proof}: We have
\begin{align*}
\|{\bf F}(\Phi x)-{\bf F}(\Phi y)\|_D =&\|\Pi(R^\pi +\gamma P^\pi\Phi x)- \Pi(R^\pi +\gamma P^\pi\Phi y) \|_D\\
\le&\|R^\pi+\gamma P^\pi\Phi x-(R^\pi+\gamma P^\pi\Phi y)\|_D\\
=& \gamma \| P^\pi\Phi(x-y)\|_D\\
\le& \gamma \|\Phi(x-y)\|_D,
\end{align*}
where the first inequality is due to the non-expansive mapping property of the projection, and the second inequality is due to~\citet[Lemma~1]{tsitsiklis1997analysis}. This completes the proof. $\quad\blacksquare$

\begin{lemma}\label{lemma:projected-Bellman-eq}
$\theta_{k+1}^*$ in~\cref{algo:deterministic-algorithm} satisfies $\Phi\theta_{k+1}^*= {\bf F}(\Phi\theta'_k)$.
\end{lemma}
{\bf Proof}: The result follows by solving the optimality condition $\nabla_\theta  l(\theta;\theta'_k)=-\Phi^T D (R^\pi+\gamma P^\pi\Phi \theta'_k-\Phi \theta)=0$. In particular, it implies
\begin{align*}
\Phi^T D\Phi\theta=\Phi^T D (R^\pi+\gamma P^\pi\Phi \theta'_k).
\end{align*}

Multiplying both sides by $(\Phi^T D\Phi)^{-1}$ from the left, we have
\begin{align*}
\theta=(\Phi^T D\Phi)^{-1}\Phi^T D (R^\pi+\gamma P^\pi\Phi \theta'_k).
\end{align*}

Again, we multiply both sides by $\Phi$ from the left to obtain
\begin{align*}
\Phi\theta=\Phi(\Phi^T D\Phi)^{-1}\Phi^T D (R^\pi+\gamma P^\pi\Phi \theta'_k)=\Pi (R^\pi+\gamma P^\pi\Phi \theta'_k).
\end{align*}
where $\Pi:=\Phi(\Phi^T D\Phi)^{-1}\Phi^T D$. This completes the proof. $\quad\blacksquare$

\begin{lemma}\label{lemma:strong-convex-f}
$l(\theta;\theta'_k):=\frac{1}{2} \|R^\pi +\gamma P^\pi \Phi\theta'_k-\Phi\theta\|_D^2$ is $\mu$-strongly convex with $\mu:=\lambda_{\min}(\Phi^T D\Phi)$.
\end{lemma}
{\bf Proof}: Noting that
\begin{align*}
&l(\theta;\theta'_k)=\frac{1}{2}(R^\pi+\gamma P^\pi\Phi\theta'_k)^T D (R^\pi+\gamma P^\pi\Phi\theta'_k)+ \frac{1}{2}\theta^T \Phi^T D \Phi\theta-(R^\pi+\gamma P^\pi\Phi\theta'_k)^T D (\Phi\theta)
\end{align*}
and that $\Phi^T D \Phi-\lambda_{\min} (\Phi^T D \Phi)I  \succeq 0$, we conclude that $l(\theta;\theta'_k)- \frac{1}{2} \|\theta\|_2^2 \lambda_{\min}(\Phi^T D \Phi)$ is convex. Therefore, by the definition of the strongly convex function, the desired conclusion holds. $\quad\blacksquare$

{\bf Proof of~\cref{prop:convergence-deterministic-GTD}}: We have
\begin{align*}
\|\Phi\theta_{k+1}-\Phi\theta^*\|_D =& \|\Phi\theta_{k+1}-\Phi\theta_{k+1}^*+\Phi\theta_{k+1}^*-\Phi\theta^*\|_D\\
\le& \| \Phi \theta_{k+1}-\Phi \theta_{k+1}^*\|_D + \|\Phi\theta_{k+1}^*- \Phi\theta^*\|_D\\
\le& \|\Phi\|_D \| \theta_{k+1}-\theta_{k+1}^*\|_D+ \|\Phi\theta_{k+1}^* -\Phi\theta^*\|_D\\
\le& \sqrt{\max_{s \in {\cal S}} d(s)} \|\Phi\|_D \sqrt {\varepsilon_{k+1}}+ \|\Phi \theta_{k+1}^*-\Phi\theta ^*\|_D\\
=& \sqrt{\max_{s \in {\cal S}} d(s)} \| \Phi \|_D \sqrt{\varepsilon_{k+1}}+ \|{\bf F}(\Phi\theta_k)-{\bf F}(\Phi\theta^*)\|_D\\
\le& \sqrt{\max_{s \in {\cal S}} d(s)}\| \Phi \|_D \sqrt{\varepsilon_{k+1}}+\gamma \|\Phi \theta_k -\Phi\theta^*\|_D,
\end{align*}
where the second equality is due to~\cref{lemma:projected-Bellman-eq} and the last inequality is due to~\cref{lemma:contraction}. Combining the last inequality over $k=0,1,\ldots,T-1$, one gets the desired result. The last result is obtained by using the Markov inequality. $\quad \blacksquare$

Note that the second term in the inequality of~\cref{prop:convergence-deterministic-GTD} vanishes as $T \to \infty$. The first terms depend on the error incurred at each iteration. In particular, if $\varepsilon_k=\varepsilon$ for all $k \geq 0$, then
\begin{align*}
&\|\Phi\theta_T-\Phi\theta^*\|_D \le\frac{\sqrt{\max_{s \in {\cal S}} d(s)} \|\Phi\|_D \sqrt \varepsilon}{1-\gamma}+ \gamma^T \| \Phi \theta_0-\Phi\theta^*\|_D.
\end{align*}
Therefore, we have
\begin{align*}
&\lim_{T\to\infty} \| \Phi \theta_T-\Phi \theta^*\|_D\le \frac{\sqrt{\max_{s \in {\cal S}} d(s)} \|\Phi\|_D\sqrt\varepsilon}{1 - \gamma}.
\end{align*}

The remaining error term can vanish if $\varepsilon \to 0$, and it can be done by increasing $L_k \to \infty$.

Finally, the proof of~\cref{prop:convergence-stochastic-GTD} follows similar lines to the proof of~\cref{prop:convergence-deterministic-GTD} except for the expectation.

{\bf Proof of~\cref{prop:convergence-stochastic-GTD}}: We have
\begin{align*}
{\mathbb E}[\|\Phi\theta_{k+1}-\Phi\theta^*\|_D] =& {\mathbb E}[\|\Phi\theta_{k+1}-\Phi\theta_{k+1}^*+\Phi\theta_{k+1}^*-\Phi\theta^*\|_D] \\
\le& {\mathbb E}[ \| \Phi \theta_{k+1}-\Phi \theta_{k+1}^*\|_D] + {\mathbb E}[ \|\Phi\theta_{k+1}^*- \Phi\theta^*\|_D]\\
\le& \|\Phi\|_D {\mathbb E}[ \|\theta_{k+1}-\theta_{k+1}^*\|_D] + {\mathbb E}[\|\Phi\theta_{k+1}^* -\Phi\theta^*\|_D]\\
\le& \sqrt{\max_{s\in {\cal S}} d(s)} \|\Phi\|_D \sqrt {\varepsilon_{k+1}}+ {\mathbb E}[\|\Phi\theta_{k+1}^*-\Phi\theta^*\|_D]\\
=& \sqrt{\max_{s\in {\cal S}} d(s)}\|\Phi\|_D \sqrt{\varepsilon_{k+1}}+ {\mathbb E}[\|{\bf F}(\Phi\theta_k)-{\bf F}(\Phi\theta^*)\|_D]\\
\le& \sqrt{\max_{s \in {\cal S}} d(s)}\|\Phi\|_D \sqrt{\varepsilon_{k+1}}+\gamma {\mathbb E}[\|\Phi\theta_k -\Phi\theta^*\|_D],
\end{align*}
where the third inequality is due to ${\mathbb E}[\sqrt{ \|\theta_{k+1}-\theta_{k+1}^*\|_2^2}]\le \sqrt{{\mathbb E}[\|\theta_{k+1}-\theta_{k+1}^* \|_2^2]}\le\sqrt{\varepsilon_{k+1}}$. Therefore, we have
\begin{align*}
&{\mathbb E}[\|\Phi\theta_{k+1}-\Phi\theta^*\|_D]\le \|\Phi\|_D \sqrt{\max_{s \in {\cal S}} d(s)}\sqrt{\varepsilon_{k+1}}+\gamma {\mathbb E}[\|\Phi\theta_k-\Phi\theta^*\|_D].
\end{align*}

Combining the last inequality over $k=0,1,\ldots,T-1$, the desired result is obtained. $\quad \blacksquare$

\vskip 0.2in

\section{Proof of~\cref{prop:convergence-SGD}}\label{app:prop_SGD_convergence}

The convergence results in~\citet[Theorem~4.7]{bottou2018optimization} can be applied to the procedure ${\tt SGD}$ of~\cref{algo:stochastic-algorithm}. We first summarize the results in~\citet{bottou2018optimization}. Consider the optimization problem
\begin{align*}
&\theta^*:=\argmin_{\theta\in {\mathbb R}^n} F(\theta),
\end{align*}
where $F:{\mathbb R}^n \to {\mathbb R}$, let $g(\theta_t)$ be an unbiased i.i.d. stochastic estimation of $\nabla_\theta F(\theta)$ at $\theta=\theta_t$, and consider the stochastic gradient descent method in~\cref{algo:stochastic-gradient-descent}.
\begin{algorithm}[h!]
\caption{Stochastic Gradient Descent (SGD)}
  \begin{algorithmic}[1]
    \State Initialize $\theta_0$.
    \For{iteration $t=0,1,\ldots$}
    	\State Compute a stochastic vector $g(\theta_t)$
        \State Choose a step size $\beta_t>0$
        \State Set the new iterate as $\theta_{t+1}=\theta_t - \beta_t g(\theta_t)$.
    \EndFor
  \end{algorithmic}\label{algo:stochastic-gradient-descent}
\end{algorithm}

With appropriate assumptions, its convergence can be proved. We first list the assumptions.
\begin{assumption}\label{assumptions-convergence}
The objective function, $F$, and SGD~\cref{algo:stochastic-gradient-descent} satisfy the following conditions:
\begin{enumerate}
\item $F$ is continuously differentiable and $\nabla_\theta F$ is Lipschitz continuous with Lipschitz constant $L<0$, i.e., $\|\nabla F(\theta)-\nabla F(\theta')\|_2 \le L \|\theta-\theta'\|_2$ for all $\theta,\theta' \in {\mathbb R}^n$.

\item $F$ is $c$-strongly convex.

\item The sequence of iterates $\{ \theta_t\}_{t=0}^\infty$ is contained in an open set over which $F$ is bounded below by a scalar $F_{\rm inf}$.

\item There exist scalars $\mu_G \geq \mu >0$ such that, for all $t \geq 0$,
\begin{align*}
&\nabla F(\theta_t)^T {\mathbb E}[g(\theta_t)|\theta_t] \ge \mu \| \nabla F(\theta_t) \|_2^2
\end{align*}
and
\begin{align*}
&\| {\mathbb E}[g(\theta_t)|\theta_t] \|_2 \le \mu_G \| \nabla F(\theta_t)\|_2.
\end{align*}

\item There exist scalars $M\geq 0$ and $M_V \geq 0$ such that, for all $k\geq 0$,
\begin{align*}
&{\mathbb V}[g(\theta_t)|\theta_t]:={\mathbb E}[\| g(\theta_t) \|_2^2|\theta_t]-\|{\mathbb E}[g(\theta_t)|\theta_t]\|_2^2 \le M + M_V \| \nabla F(\theta_t)\|_2^2
\end{align*}
\end{enumerate}
\end{assumption}

Under~\cref{assumptions-convergence}, the convergence of iterates of~\cref{algo:stochastic-gradient-descent} in expectation can be proved.
\begin{lemma}\label{lemma:stochastic-gradient-convergence}
Under~\cref{assumptions-convergence} (with $F_{\inf}=F(\theta^*)$), suppose that the SGD method in~\cref{algo:stochastic-gradient-descent} is run with a stepsize sequence such that, for all $t\geq 0$,
\begin{align*}
\beta_t=\frac{\beta}{\kappa+t+1}
\end{align*}
for some $\beta>1/(c\mu)$ and $\kappa >0$ such that $\beta_0 \le \mu/(L(M+\mu_G^2))$. Then, for all $t \geq 0$, the expected optimality gap satisfies
\begin{align*}
{\mathbb E}[F(\theta_t)-F(\theta^*)] \le \frac{\nu}{\kappa+t+1},
\end{align*}
where
\begin{align*}
&\nu:=\max \left\{ \frac{\beta^2 LM}{2(\beta c\mu-1)},(\kappa+1)(F(\theta_0)-F(\theta^*)) \right\}
\end{align*}
\end{lemma}

To apply~\cref{lemma:stochastic-gradient-convergence} to $\tt SGD$ of~\cref{algo:stochastic-algorithm}, we will prove that all the conditions in~\cref{assumptions-convergence} are satisfied with $F(\theta)=l(\theta;\theta'_k):=\frac{1}{2} \|R^\pi +\gamma P^\pi \Phi\theta'_k-\Phi\theta\|_D^2$. The strong convexity is established in~\cref{lemma:strong-convex-f}. In the following lemmas, we prove the Lipschitz continuity of the gradient and the remaining conditions in~\cref{assumptions-convergence}.

\begin{lemma}[Lipschitz continuous gradient]
$F$ satisfies
\begin{align*}
\|\nabla F(\theta)-\nabla F(\theta')\|_2 \le L \|\theta-\theta'\|_2,\quad \forall \theta,\theta' \in {\mathbb R}^n
\end{align*}
with $L = \sqrt{\lambda_{\max} (\Phi^T D\Phi \Phi^T D\Phi)}$.
\end{lemma}
{\bf Proof}: Noting that $\nabla F(\theta)=\Phi^T D(R^\pi+P^\pi \Phi\theta_k-\Phi \theta)$, we have
\begin{align*}
\| \nabla F(\theta)-\nabla F(\theta') \|_2=&\|\Phi^T D(R^\pi+P^\pi \Phi\theta_k -\Phi \theta)-\Phi^T D(R^\pi+P^\pi\Phi\theta_k -\Phi \theta')\|_2\\
=&\|\Phi^T D\Phi(\theta - \theta')\|_2\\
\le& \sqrt{\lambda_{\max} (\Phi^T D\Phi \Phi^T D\Phi)} \|\theta-\theta'\|_2,
\end{align*}
which proves the desired result. $\quad\blacksquare$

\begin{lemma}\label{lemma:bounds-on-gradient}
For $\tt SGD$ of~\cref{algo:stochastic-algorithm}, we have
\begin{align*}
{\mathbb E}[g(\theta_{k,t})|\theta_{k,t},\theta_k] =& \left. \nabla_\theta \left(\frac{1}{2}\| R^\pi+\gamma P^\pi  \Phi\theta_k-\Phi\theta\|_D^2 \right)\right|_{\theta=\theta_{k,t}},\\
{\mathbb E}[\| g(\theta_{k,t}) \|_2^2|\theta_{k,t},\theta_k] \le& \|\Phi\|_2^2 (3\sigma^2+3\|\Phi\|_2^2 \|\theta_k\|_2^2 +3\|\Phi\|_2^2 \|\theta_{k,t}\|_2^2).
\end{align*}
\end{lemma}
{\bf Proof}: By the definition of $g(\theta_{k,t})$ in $\tt SGD$ of~\cref{algo:stochastic-algorithm}, we have
\begin{align*}
{\mathbb E}[ g(\theta_{k,t})|\theta_{k,t},\theta_k]=&{\mathbb E}[\phi(s)(r^\pi(s) + \phi^T(s')\theta_k -\phi(s)^T \theta_{k,t})|\theta_{k,t},\theta_k]\\
=& {\mathbb E}[\Phi^T e_s(r^\pi(s)+e_{s'}^T \Phi\theta_k -e_s^T \Phi \theta_{k,t})|\theta_{k,t},\theta_k]\\
=& {\mathbb E}[\Phi^T e_s e_s^T(R^\pi +e_s e_{s'}^T \Phi\theta_k -\Phi\theta_{k,t})|\theta_{k,t},\theta_k]\\
=&\Phi^T D(R^\pi+P^\pi\Phi\theta_k-\Phi\theta_{k,t})\\
=& \left. \nabla_\theta \left( \frac{1}{2}\|R^\pi +P^\pi\Phi\theta_k-\Phi\theta\|_D^2\right) \right|_{\theta= \theta_{k,t}},
\end{align*}
proving the first equation. For the second result, we have
\begin{align*}
\| g(\theta_{k,t}) \|_2^2 =& \| \Phi^T e_s(r^\pi(s)+e_{s'}^T \Phi\theta_k-e_s^T\Phi\theta_{k,t})\|_2^2\\
\le& \|\Phi\|_2^2 \|r^\pi(s)+e_{s'}^T \Phi\theta_k-e_s^T \Phi\theta_{k,t}\|_2^2\\
\le& \|\Phi\|_2^2(3\|r^\pi(s)\|_2^2+3\|e_{s'}^T\Phi\theta_k\|_2^2 +3\|e_s^T \Phi\theta_{k,t}\|_2^2 )\\
\le& \|\Phi\|_2^2 (3\sigma^2+3\|\Phi\|_2^2 \|\theta_k\|_2^2 +3\|\Phi\|_2^2 \|\theta_{k,t}\|_2^2),
\end{align*}
proving the second result. $\quad\blacksquare$

The first result in~\cref{lemma:bounds-on-gradient} implies that $g(\theta_{k,t})$ is an unbiased stochastic estimation of $\nabla F(\theta_{k,t})$. The second result in~\cref{lemma:bounds-on-gradient} means that the second moment of the stochastic gradient estimation is bounded by a quantity which is dependent on $\|\theta_k\|_2^2$. Based on~\cref{lemma:bounds-on-gradient}, we bound the variance of the gradient in the next lemma. Before proceeding, we introduce an inequality which will be frequently used.
\begin{lemma}\label{lemma:inequalities}
For any $a,b \in {\mathbb R}^n$, we have
\begin{align*}
&\|a+b\|_2^2 \le (1+\varepsilon)\|a\|_2^2 + (1+\varepsilon^{-1})\|b\|_2^2,\\
&\|a+b\|_2^2 \ge (1-\varepsilon)\|a\|_2^2 +(1-\varepsilon^{-1})\|b\|_2^2
\end{align*}
where $\varepsilon >0$ is any real number.
\end{lemma}
{\bf Proof}: We obtain the first upper bound by
\begin{align*}
\|a+b\|_2^2=&\|a\|_2^2 + \|b\|_2^2 + 2a^T b\\
\le& \|a\|_2^2 + \|b\|_2^2 + 2|a^T b|\\
\le& \|a\|_2^2 + \|b\|_2^2 + \varepsilon \|a\|_2^2 + \varepsilon^{-1} \| b\|_2^2
\end{align*}
for any $\varepsilon>0$, where the last inequality is due to the Young's inequality, $|a^T b| \le \varepsilon \|a\|_2^2/2 +\varepsilon^{-1} \| b\|_2^2/2$. Similarly, the lower bound can be obtained by
\begin{align*}
\|a+b\|_2^2=&\|a\|_2^2 + \|b\|_2^2 + 2a^T b\\
\ge& \|a\|_2^2 + \|b\|_2^2 -2|a^T b|\\
\ge& \|a\|_2^2 +\|b\|_2^2 -\varepsilon \|a\|_2^2 -\varepsilon^{-1}\|b\|_2^2.
\end{align*}
This completes the proof. $\quad\blacksquare$

\begin{lemma}[Bounded variance]\label{lemma:bounded-variance}
The variance of the gradient is bounded as follows:
\begin{align*}
&{\mathbb V}[g(\theta_{k,t})|\theta_{k,t},\theta_k] \le \xi_1 + \xi_2 \| \theta_k \|_2^2 +\xi_3 \| \nabla F(\theta_{k,t}) \|_2^2,
\end{align*}
where
\begin{align*}
&\xi_1:=3\sigma^2 \|\Phi\|_2^2 + 2(1+\xi_3)^2\| \Phi^T DR^\pi \|_2^2\\
&\xi_2:=3\| \Phi \|_2^4 +2(1+\xi_3)^2\lambda_{\max}(\Phi^T(P^\pi)^T D\Phi\Phi^T DP^\pi \Phi)\\
&\xi_3:=\frac{3\|\Phi\|_2^4}{\lambda_{\min}(\Phi^T D\Phi\Phi^T D\Phi)}.
\end{align*}
\end{lemma}
{\bf Proof}: Using the definition of ${\mathbb V}[g(\theta_t)|\theta_{k,t},\theta_k]$ in~\cref{assumptions-convergence} and the bound on ${\mathbb E}[\| g(\theta_{k,t})\|_2^2|\theta_{k,t},\theta_k]$ in~\cref{lemma:bounds-on-gradient}, we have
\begin{align}
{\mathbb V}[g(\theta_{k,t})|\theta_{k,t},\theta_k] &= {\mathbb E}[\| g(\theta_{k,t})\|_2^2|\theta_{k,t},\theta_k]-\| {\mathbb E}[g(\theta_{k,t})|\theta_{k,t},\theta_k]\|_2^2\nonumber\\
&= {\mathbb E}[\|g(\theta_{k,t})\|_2^2|\theta_{k,t},\theta_k] - \| \nabla F(\theta_{k,t})\|_2^2\nonumber\\
&= {\mathbb E}[\|g(\theta_{k,t})\|_2^2|\theta_{k,t},\theta_k]-(1+K) \|\nabla F(\theta_{k,t})\|_2^2 +K\|\nabla F(\theta_{k,t})\|_2^2\nonumber\\
&\le \|\Phi\|_2^2(3\sigma^2 +3\|\Phi\|_2^2 \|\theta_k\|_2^2 + 3\|\Phi\|_2^2 \|\theta_{k,t}\|_2^2) - (1+K)\|\nabla F(\theta_{k,t})\|_2^2  + K\| \nabla F(\theta_{k,t})\|_2^2\label{eq:6}
\end{align}
for any $K>0$. A main issue in~\eqref{eq:6} is the presence of the term depending on $\theta_{k,t}$. We will obtain a bound on the first two terms which does not depend on $\theta_{k,t}$. To this end, a lower bound on $\| \nabla F(\theta_{k,t})\|_2^2$ is obtained as follows:
\begin{align*}
\|\nabla F(\theta_{k,t})\|_2^2 =&\|\Phi^T DR^\pi+\Phi^TDP^\pi\Phi\theta_k-\Phi^TD\Phi\theta_{k,t}\|_2^2\\
\ge& (1-\varepsilon^{-1})\|\Phi^T D\Phi\theta_{k,t}\|_2^2 +(1-\varepsilon)\| \Phi^T DR^\pi+\Phi^T DP^\pi\Phi\theta_k \|_2^2\\
\ge& (1-\varepsilon^{-1})\lambda_{\min}(\Phi^T D\Phi\Phi^T D\Phi) \|\theta_{k,t}\|_2^2 - (1-\varepsilon)\| \Phi^TDR^\pi + \Phi^TDP^\pi\Phi\theta_k \|_2^2,
\end{align*}
for any $\varepsilon>0$ such that $1-\varepsilon^{-1}>0$, where the first inequality is due to~\cref{lemma:inequalities}. Combining the last inequality with~\eqref{eq:6} yields
\begin{align}
{\mathbb V}[g(\theta_{k,t})|\theta_{k,t},\theta_k]\le& 3\sigma^2 \|\Phi\|_2^2 + 3\|\Phi\|_2^4 \|\theta_k\|_2^2  + \{3\|\Phi\|_2^4 -(1+K)(1-\varepsilon^{-1})\lambda_{\min}(\Phi^T D\Phi\Phi^T D\Phi)\} \| \theta_{k,t}\|_2^2\nonumber\\
&-(1+K)(1-\varepsilon) \|\Phi^T DR^\pi +\Phi^T DP^\pi \Phi\theta_k \|_2^2 +K\|\nabla F(\theta_{k,t})\|_2^2.\label{eq:7}
\end{align}

Note that by appropriately choosing $\varepsilon>0$ and $K>0$, the term related to $\|\theta_{k,t}\|_2^2$ can be removed. In particular, we can choose $\varepsilon>0$ and $K>0$ such that $3\|\Phi\|_2^4 -(1+K)(1-\varepsilon^{-1})\lambda_{\min} (\Phi^T D\Phi\Phi^T D\Phi)=0$. A solution is
\begin{align*}
\varepsilon =\frac{\delta\lambda_{\min}(\Phi^T D\Phi\Phi^T D\Phi)+3\|\Phi\|_2^4}{\delta\lambda_{\min}(\Phi^T D\Phi\Phi^T D\Phi)}
\end{align*}
and
\begin{align*}
K=\frac{3 \|\Phi\|_2^4}{\lambda_{\min}(\Phi^T D\Phi\Phi^T D\Phi)}-1+\delta
\end{align*}
for any $\delta>0$. Setting $\delta =1$ and substituting these expressions for $\varepsilon$ and $K$ in~\eqref{eq:7} result in
\begin{align}
{\mathbb V}[g(\theta_{k,t})]\le& 3\sigma^2 \|\Phi\|_2^2 +3\|\Phi\|_2^4 \|\theta _k\|_2^2  + (1+\xi_3)\xi_3\|\Phi^T DR^\pi +\Phi^T DP^\pi\Phi\theta_k \|_2^2 + K \|\nabla F(\theta_{k,t})\|_2^2\nonumber\\
\le& 3\sigma^2 \|\Phi\|_2^2 +3\|\Phi\|_2^4 \|\theta_k\|_2^2 + (1+\xi_3)^2 \|\Phi^T DR^\pi +\Phi^T DP^\pi\Phi\theta_k \|_2^2 + K \|\nabla F(\theta_{k,t})\|_2^2\label{eq:14}
\end{align}
where
\begin{align*}
&\xi_3:=\frac{3\|\Phi\|_2^4}{\lambda_{\min}(\Phi^T D\Phi\Phi^T D\Phi)}.
\end{align*}

Applying~\cref{lemma:inequalities} again for $\|\Phi^T DR^\pi +\Phi^T DP^\pi\Phi\theta_k \|_2^2$ in~\eqref{eq:14} yields
\begin{align*}
&{\mathbb V}[g(\theta_{k,t})|\theta_{k,t},\theta_k] \le\xi_1 +\xi_2 \| \theta_k \|_2^2 +\xi_3 \| \nabla F(\theta_{k,t}) \|_2^2,
\end{align*}
where
\begin{align*}
&\xi_1:=3\sigma^2 \|\Phi\|_2^2 + 2(1+\xi_3)^2 \| \Phi^T DR^\pi \|_2^2\\
&\xi_2:=3\| \Phi \|_2^4 +2(1+\xi_3)^2\lambda_{\max}(\Phi^T(P^\pi)^T D\Phi\Phi^T DP^\pi \Phi)
\end{align*}
which is the desired conclusion. $\quad\blacksquare$

We are now ready to prove~\cref{prop:convergence-SGD}.

{\bf Proof of~\cref{prop:convergence-SGD}}: The first statement of~\cref{prop:convergence-SGD} is proven in~\cref{lemma:bounded-variance}. To prove the remaining conditions of~\cref{prop:convergence-SGD}, we note that all the results of this section prove that~\cref{assumptions-convergence} is satisfied with $\mu=\mu_G=1,c =\lambda_{\min}(\Phi^T D\Phi),L=\sqrt{\lambda_{\max}(\Phi^T D\Phi\Phi^T D\Phi)},M=\xi_1+\xi_2 \|\theta_k\|_2^2$, $M_V=\xi_3$, and $F_{\rm inf} = \min_{\theta} F(\theta)$, where the positive real numbers $\xi_1$, $\xi_2$, and $\xi_3$ are given in~\cref{lemma:bounded-variance}. Then, by using~\cref{lemma:stochastic-gradient-convergence}, it can be proved that if the SGD method in~\cref{algo:stochastic-algorithm} is run with a stepsize sequence such that, for all $t\geq 0$,
\begin{align*}
\beta_t=\frac{\beta}{\kappa+t+1}
\end{align*}
for some $\beta>1/\lambda_{\min}(\Phi^T D\Phi)$ and $\kappa >0$ such that
\begin{align*}
&\beta_0 =\frac{\beta}{\kappa+1}\le \frac{1}{\sqrt{\lambda_{\max}(\Phi^T D\Phi\Phi^T D\Phi)}(\xi_3+1)},
\end{align*}
then, for all $0\leq t\leq L_k-1$, the expected optimality gap satisfies
\begin{align}
{\mathbb E}[F(\theta_{k,t})-F(\theta_{k+1}^*)|\theta_k] \le \frac{\nu}{\kappa+t+1}\label{eq:8}
\end{align}
with
\begin{align*}
\nu=\max\left\{\frac{\beta^2\sqrt{\lambda_{\max}(\Phi^T D\Phi\Phi^T D\Phi)}(\xi_1+\xi_2 \| \theta_k \|_2^2)}{2(\beta \lambda_{\min} (\Phi^TD\Phi)-1)},(\kappa+1)(F(\theta_k)-F(\theta_{k+1}^*)) \right\}.
\end{align*}

Note that $\theta_{k+1}^*$ is the solution that minimizes $F$, which is different from $\theta^*$. By the definition of the strong convexity, we have
\begin{align*}
&F(x)+\nabla F(x)^T (y-x)+\frac{\lambda_{\min}(\Phi^TD\Phi)}{2} \|x-y\|_2^2 \le F(y),\quad \forall x,y \in {\mathbb R}^n.
\end{align*}

Letting $x=\theta_{k+1}^*,y=\theta_{k,t}$ in the above inequality yields
\begin{align*}
\frac{\lambda_{\min}(\Phi^TD\Phi)}{2} \|\theta_{k+1}^*-\theta_{k,t}\|_2^2\le F(\theta_{k,t})-F(\theta_{k+1}^*),
\end{align*}
where we use the fact that $\theta_{k+1}^*$ minimizes $F$. Combining the last inequality with~\eqref{eq:8}, we get
\begin{align}
&{\mathbb E}[\|\theta_{k+1}^*-\theta_{k,t}\|_2^2|\theta_k]\le \frac{2}{\lambda_{\min}(\Phi^TD\Phi)}\frac{\nu}{\kappa+t+1}.\label{eq:10}
\end{align}

For later analysis, we will further polish the upper bound. Using $F(\theta_{k+1}^*) \geq 0$ and the triangle inequality, we have
\begin{align}
\nu&=\max \left\{ \frac{\beta^2\sqrt{\lambda_{\max} (\Phi^T D\Phi\Phi^T D\Phi)} (\xi_1+\xi_2\|\theta_k\|_2^2)}{2(\beta \lambda _{\min } (\Phi^TD\Phi)-1)},(\kappa+1)(F(\theta_k)-F(\theta_{k+1}^*)) \right\}\nonumber\\
&\le\max \left\{ \frac{\beta^2\sqrt{\lambda_{\max}(\Phi^T D\Phi\Phi^TD\Phi)} \left(\xi_1+\xi_2 \|\theta^*\|_2^2 +\xi_2 \|\theta_k -\theta^*\|_2^2 \right)}{2(\beta\lambda_{\min}(\Phi^T D\Phi)-1)},(\kappa+1)F(\theta_k)\right\},\label{eq:9}
\end{align}
where $\theta^*$ is the solution of the projected Bellman equation, $\Phi \theta^*= {\bf F}(\Phi\theta^*)$, that we want to find, and it should not be confused with $\theta_{k+1}^*$, which is the solution that minimizes $F$.

Next, $F(\theta_k)$ is bounded as
\begin{align*}
F(\theta_k)&=\frac{1}{2} \|R^\pi+P^\pi \Phi\theta_k -\Phi\theta_k\|_D^2\\
&= \frac{1}{2} \|R^\pi+P^\pi \Phi\theta_k -\Phi\theta_k -(R^\pi +P^\pi \Phi\theta^* -\Phi\theta^*)+(R^\pi +P^\pi \Phi\theta^*-\Phi\theta^*)\|_D^2\\
&\le \|R^\pi+P^\pi\Phi\theta_k-\Phi\theta_k-(R^\pi+P^\pi \Phi\theta^* -\Phi\theta^*)\|_D^2+\|R^\pi+P^\pi\Phi\theta^*-\Phi\theta^*\|_D^2\\
&= \| (P^\pi\Phi-\Phi)(\theta_k-\theta^*)\|_D^2 + \| R^\pi +P^\pi \Phi\theta^*-\Phi\theta^*\|_D^2\\
&\le \lambda_{\max}((P^\pi\Phi-\Phi)^T D(P^\pi\Phi-\Phi)) \| \theta_k-\theta^* \|_2^2 + \|R^\pi +P^\pi \Phi\theta^*-\Phi\theta^*\|_D^2,
\end{align*}
where the first inequality is due to the inequality $(a+b)^2 \leq 2a^2 +2b^2$. We combine this result with~\eqref{eq:9} to obtain
\begin{align*}
\nu\le& \max \left\{ \frac{\beta^2 \sqrt{\lambda_{\max}(\Phi^T D\Phi\Phi^T D\Phi)} (\xi_1 + \xi_2 \| \theta^*\|_2^2 +\xi_2 \| \theta_k -\theta^* \|_2^2)}{2(\beta\lambda_{\min}(\Phi^T D\Phi)-1)},\right.\\
& \left. (\kappa+1)\lambda_{\max}((P^\pi\Phi-\Phi)^T D(P^\pi\Phi-\Phi )) \| \theta_k-\theta^* \|_2^2 + (\gamma+1)\| R^\pi+P^\pi \Phi\theta^*-\Phi\theta^*\|_D^2 \right\}\\
\le& \frac{\beta^2 \sqrt{\lambda_{\max} (\Phi^T D\Phi\Phi^T D\Phi)} \left(\xi_1+\xi_2\|\theta^*\|_2^2+\xi_2 \|\theta_k-\theta^*\|_2^2 \right)}{2(\beta\lambda_{\min}(\Phi^TD\Phi)-1)}\\
&+ (\kappa+1) \lambda_{\max}((P^\pi\Phi-\Phi)^T D(P^\pi\Phi-\Phi)) \|\theta_k-\theta^*\|_2^2 + (\kappa+1) \|R^\pi +P^\pi\Phi\theta^*-\Phi\theta^*\|_D^2\\
=& \chi_1+\chi_2 \|\theta_k-\theta^*\|_2^2,
\end{align*}
where the second inequality is due to the inequality $\max \{a,b\}\le a+b$ and
\begin{align*}
&\chi_1=\frac{\beta^2\sqrt{\lambda_{\max}(\Phi^TD\Phi\Phi^TD\Phi)}(\xi_1 +\xi_2 \| \theta^*\|_2^2)}{2(\beta\lambda_{\min}(\Phi^T D\Phi)-1)} + (\kappa+1)\| R^\pi+P^\pi\Phi\theta^*-\Phi\theta^*\|_D^2,\\
&\chi_2=\frac{\beta^2\xi_2\sqrt{\lambda_{\max}(\Phi^T D\Phi\Phi^T D\Phi)}}{2(\beta\lambda_{\min} (\Phi^TD\Phi)-1)}+ (\kappa+1)\lambda_{\max}((P^\pi\Phi-\Phi)^T D(P^\pi\Phi-\Phi)).
\end{align*}

Plugging the upper bound into $\nu$ in~\eqref{eq:10} and after simplifications, the desired result follows. $\quad\blacksquare$

\vskip 0.2in
\section{Proof of~\cref{prop:sample-complexity}}\label{app:prop_sample_complexity}

To prove the sample complexity, we will make use of~\cref{prop:convergence-SGD}. A tricky part is due to the fact that the constant factor of the convergence rate depends on $\|\theta_k-\theta^*\|_2^2$. We will prove that $\|\theta_k-\theta^*\|_2^2$ is bounded by a constant in expectation, which plays a key role in the proof.
\begin{lemma}\label{lemma:bounded-second-moment}
Suppose that~\cref{algo:stochastic-algorithm} is run with $\varepsilon_i=\varepsilon$ for all $k\ge i \ge 1$. Then,
\begin{align*}
&{\mathbb E}[\|\theta_i-\theta^*\|_2^2] \le \omega_1\varepsilon+\omega_2,\quad\forall 0 \leq i \leq k,
\end{align*}
where
\begin{align*}
&\omega_1:=\frac{2\left(\frac{1+\gamma^2}{1-\gamma^2} \right) \|\Phi\|_D^2 \lambda_{\max}(D)}{\lambda_{\min} (\Phi^T D\Phi)(1-\gamma^2)},\quad\omega_2:=\frac{{\mathbb E}[\|\Phi\theta_0-\Phi\theta^*\|_D^2]}{\lambda_{\min}(\Phi^TD\Phi)}.
\end{align*}
\end{lemma}
{\bf Proof}: We follow the procedure similar to that of~\cref{prop:convergence-stochastic-GTD}. The main difference relies on the fact that we need a bound on the squared norm. First, we obtain the chain of inequalities
\begin{align*}
&{\mathbb E}[\| \Phi\theta_{i+1}-\Phi\theta^*\|_D^2 |\theta_i]\\
=&{\mathbb E}[\|\Phi\theta_{i+1}-\Phi\theta_{i+1}^*+\Phi\theta_{i+1}^*-\Phi\theta^*\|_D^2 |\theta_i]\\
\leq& (1+\delta^{-1}) {\mathbb E}[\|\Phi\theta_{i+1}-\Phi\theta_{i+1}^*\|_D^2|\theta_i]+(1+\delta){\mathbb E}[\|\Phi\theta_{i+1}^*-\Phi\theta^*\|_D^2|\theta_i]\\
\le& (1+\delta^{-1}){\mathbb E}[\|\Phi\|_D^2 \|\theta_{i+1}-\theta_{i+1}^*\|_D^2|\theta_i]+(1+\delta) {\mathbb E}[\|\Phi\theta_{i+1}^*-\Phi\theta^*\|_D^2|\theta_i]\\
\le& (1+\delta^{-1}) \|\Phi\|_D^2\lambda_{\max}(D) {\mathbb E}[\| \theta_{i+1}-\theta_{i+1}^*\|_2^2|\theta_i]+(1+\delta){\mathbb E}[\|\Phi\theta_{i+1}^*-\Phi\theta^*\|_D^2 |\theta_i]\\
=&(1+\delta^{-1}) \|\Phi\|_D^2 \lambda_{\max}(D){\mathbb E}[\| \theta_{i+1}-\theta_{i+1}^*\|_2^2|\theta_i] +(1+\delta){\mathbb E}[\|{\bf F}(\Phi\theta_i)-{\bf F}(\Phi\theta^*)\|_D^2 |\theta_i]\\
\le& (1+\delta^{-1})\|\Phi\|_D^2 \lambda_{\max}(D){\mathbb E}[\| \theta_{i+1}-\theta_{i+1}^*\|_2^2|\theta_i] +(1+\delta)\gamma^2 \|\Phi\theta_i-\Phi\theta^*\|_D^2,
\end{align*}
where $0 \leq i \leq k-1$, the first equality is due to~\cref{lemma:inequalities}, the second equality follows from~\cref{lemma:projected-Bellman-eq}, and the last inequality is due to~\cref{lemma:contraction}. Since $\gamma \in [0,1)$, there exists $\delta>0$ such that $(1+\delta)\gamma^2<1$, which is equivalent to $\delta <\frac{1-\gamma^2}{\gamma^2}$. We simply choose $\delta=\frac{1-\gamma^2}{2\gamma^2}$, yielding
\begin{align*}
&{\mathbb E}[\|\Phi\theta_{i+1}-\Phi\theta^*\|_D^2|\theta_i]\le \left(1+\frac{2\gamma^2}{1-\gamma^2}\right) \|\Phi\|_D^2 \lambda_{\max} (D){\mathbb E}[\| \theta_{i+1}-\theta_{i+1}^*\|_2^2|\theta_i] + \frac{\gamma^2+1}{2}\| \Phi\theta_i-\Phi\theta^* \|_D^2.
\end{align*}

Taking the total expectation on both sides and using the hypothesis, ${\mathbb E}[\| \theta_{i+1}-\theta_{i+1}^*\|_2^2] \leq \varepsilon$ for all $0 \leq i \leq k-1$, yield
\begin{align*}
&{\mathbb E}[\|\Phi\theta_{i+1}-\Phi\theta^*\|_D^2]\le \left(1+\frac{2\gamma^2}{1-\gamma^2}\right) \|\Phi\|_D^2 \lambda_{\max} (D)\varepsilon + \frac{\gamma^2+1}{2} {\mathbb E}[\|\Phi\theta_i-\Phi\theta^*\|_D^2],\quad \forall  0 \leq i \leq k-1.
\end{align*}

By the induction argument in $i$, we have
\begin{align*}
{\mathbb E}[\|\Phi\theta_i-\Phi\theta^*\|_D^2] &\le\left(1+\frac{2\gamma^2}{1-\gamma^2} \right) \|\Phi\|_D^2 \lambda_{\max}(D)\varepsilon \sum_{t=0}^{i-1}{\left(\frac{\gamma^2+1}{2}\right)^t} + \left(\frac{\gamma^2+1}{2}\right)^i {\mathbb E}[\|\Phi\theta_0-\Phi\theta^*\|_D^2]\\
\le& \left(1+\frac{2\gamma^2}{1-\gamma^2}\right) \|\Phi\|_D^2 \lambda_{\max}(D)\varepsilon \frac{1}{1-\frac{\gamma^2+1}{2}} + \left( \frac{\gamma^2+1}{2} \right)^i {\mathbb E}[\|\Phi\theta_0-\Phi\theta^*\|_D^2]\\
\le& \left(1+\frac{2\gamma^2}{1-\gamma^2}\right) \|\Phi\|_D^2 \lambda_{\max}(D)\varepsilon \frac{1}{1-\frac{\gamma^2+1}{2}} + {\mathbb E}[\|\Phi\theta_0-\Phi\theta^*\|_D^2],\quad \forall 1 \leq i \leq k ,
\end{align*}
where the second inequality is obtained by letting $i \to \infty$ and the last inequality is due to $\left( \frac{\gamma^2+1}{2} \right)^i <1$. Since the first term on the right hand side is nonnegative, the last inequality holds for $i=0$. By using ${\mathbb E}[\|\Phi\theta_k-\Phi\theta^*\|_D^2]\ge\lambda_{\min}(\Phi^T D\Phi) {\mathbb E}[\|\theta_k-\theta^*\|_2^2]$
and arranging terms, we arrive at the conclusion. $\quad\blacksquare$

\cref{lemma:bounded-second-moment} states that if the subproblems are solved such that ${\mathbb E}[\| \theta_{i}-\theta_{i}^*\|_2^2] \leq \varepsilon$ for $k\geq i\geq 1$, then ${\mathbb E}[\|\theta_{i}-\theta^* \|_2^2]$ is bounded by a constant depending on $\varepsilon$ for all $k\geq i\geq 0$. Using this property, we introduce another version of~\cref{prop:convergence-SGD} which drops the dependency of $\|\theta_k -\theta^*\|_2^2$.
\begin{proposition}\label{prop:convergence-SGD2}
Suppose that the SGD method in~\cref{algo:stochastic-algorithm} is run with a stepsize sequence such that, for all $t\geq 0$,
\begin{align*}
\beta_t=\frac{\beta}{\kappa+t+1}
\end{align*}
for some $\beta>1/\lambda_{\min}(\Phi^T D\Phi)$ and $\kappa >0$ such that
\begin{align*}
&\beta_0 =\frac{\beta}{\kappa+2}\le \frac{1}{\sqrt{\lambda_{\max}(\Phi^T D\Phi\Phi^T D\Phi)}(\xi_3+1)}.
\end{align*}

Moreover, suppose that~\cref{algo:stochastic-algorithm} is run with $\varepsilon_i=\varepsilon$ for all $k-1 \ge i \ge 1$. Then, for all $0\leq t\leq L_k-1$, the expected optimality gap satisfies
\begin{align*}
{\mathbb E}[\|\theta_{k+1}^*-\theta_{k,t}\|_2^2]\le \frac{2}{\lambda_{\min}(\Phi^T D\Phi)}\frac{\chi_1+\chi_2( \omega_1\varepsilon+\omega_2)}{\kappa+t+1}.
\end{align*}
\end{proposition}
{\bf Proof}: The proof is completed by taking the total expectation on both sides of~\eqref{eq:12} in~\cref{prop:convergence-SGD} and using the bound in~\cref{lemma:bounded-second-moment}. $\quad\blacksquare$

From~\cref{prop:convergence-SGD2}, we concludes that with the number of subproblem iterations such that
\begin{align*}
\frac{2(\chi_1+\chi_2(\omega_1\varepsilon+\omega_2))}{\lambda_{\min}(\Phi^T D\Phi)\varepsilon}-\kappa-1\le L_k,
\end{align*}
each subproblem achieves $\varepsilon$-optimality in expectation. Based on this observation, we will now prove the sample complexity. By~\cref{prop:convergence-stochastic-GTD}, ${\mathbb E}[\|\theta_{T+1}-\theta^*\|_D] \le \epsilon$ holds if
\begin{align}
&\|\Phi\|_D \max_{s\in {\cal S}} d(s)\frac{\sqrt\varepsilon}{1-\gamma}+\gamma^T {\mathbb E}[\|\Phi\theta_0-\Phi\theta^* \|_D]\le\epsilon.\label{eq:5}
\end{align}

Again, it holds if
\begin{align}
&\|\Phi\|_D \max_{s\in {\cal {\cal S}}} d(s)\frac{\sqrt\varepsilon}{1-\gamma} \le a\epsilon\label{eq:1}
\end{align}
and
\begin{align}
&\gamma^T {\mathbb E}[\|\Phi\theta_0-\Phi\theta^*\|_D] \le b\epsilon.\label{eq:2}
\end{align}
for any real numbers $a,b > 0$ such that $a+b=1$. The condition~\eqref{eq:2} holds if
\begin{align}
&T \ge \ln\left(\frac{b\epsilon}{{\mathbb E}[\|\Phi\theta_0-\Phi\theta^*\|_D ]}\right)/\ln \gamma.\label{eq:3}
\end{align}

The condition~\eqref{eq:1} holds if
\begin{align}
&\varepsilon\le\frac{a^2 \epsilon^2(1-\gamma)^2}{\|\Phi\|_D^2 \max_{s\in {\cal S}} d(s)}.\label{eq:11}
\end{align}

Combined with~\cref{prop:convergence-SGD} and~\cref{lemma:bounded-second-moment}, a sufficient condition of~\eqref{eq:11} is
\begin{align*}
&\frac{2}{\lambda_{\min}(\Phi^TD\Phi)}\frac{\chi_1+\chi_2 \omega_1\varepsilon +\chi_2\omega_2}{\kappa+L_k+1}\le \frac{a^2\epsilon^2(1-\gamma)^2}{\|\Phi\|_D^2 \max_{s\in {\cal S}} d(s)}.
\end{align*}

Using the upper bound in~\eqref{eq:11} and arranging terms, we have that the condition~\eqref{eq:11} (and hance~\eqref{eq:1}) holds if the number of iteration, $L_k$, for the subproblem at iteration $k$ is lower bounded by
\begin{align}
&\frac{2(\chi_1+\chi_2\omega_2)}{\lambda_{\min}(\Phi^T D\Phi)}\frac{\|\Phi\|_D^2 \max_{s\in {\cal S}} d(s)}{a^2 \epsilon^2 (1-\gamma)^2}+\frac{2\chi_2\omega_1}{\lambda_{\min}(\Phi^T D\Phi)}-\kappa-1 \le L_k\label{eq:4}
\end{align}

Combining~\eqref{eq:3} and~\eqref{eq:4}, we conclude that~\eqref{eq:5} holds with SO calls at most
\begin{align*}
&\frac{1}{\ln\gamma^{-1}}\left\{ \frac{2(\chi_1+\chi_2\omega_2)}{\lambda_{\min}(\Phi^TD\Phi)}\frac{\|\Phi\|_D^2\max_{s\in {\cal S}}d(s)}{a^2\epsilon^2(1-\gamma)^2} + \frac{2\chi_2\omega_1}{\lambda_{\min}(\Phi^TD\Phi)}-\kappa-1\right\}\left\{\ln\left(\frac{{\mathbb E}[\|\Phi\theta_0-\Phi\theta^*\|_D]}{b \epsilon }\right) \right\}.
\end{align*}

To simplify the expression, $a >0$ and $b>0$ are set to be $a=\sqrt{\max_{s\in {\cal S}} d(s)}$ and $b=1-\sqrt{\max_{s\in {\cal S}} d(s)}$, respectively. Plugging the explicit expressions  for $\omega_1,\omega_2$ in~\cref{lemma:bounded-second-moment} and further simplifications lead to the desired conclusion.

\vskip 0.2in
\section{Additional Simulations for~\cref{section:simulations}}\label{app:additional-simulations}
\begin{figure*}[h!]
\centering\subfigure[{Standard TD}]{\includegraphics[width=8cm,height=6cm]{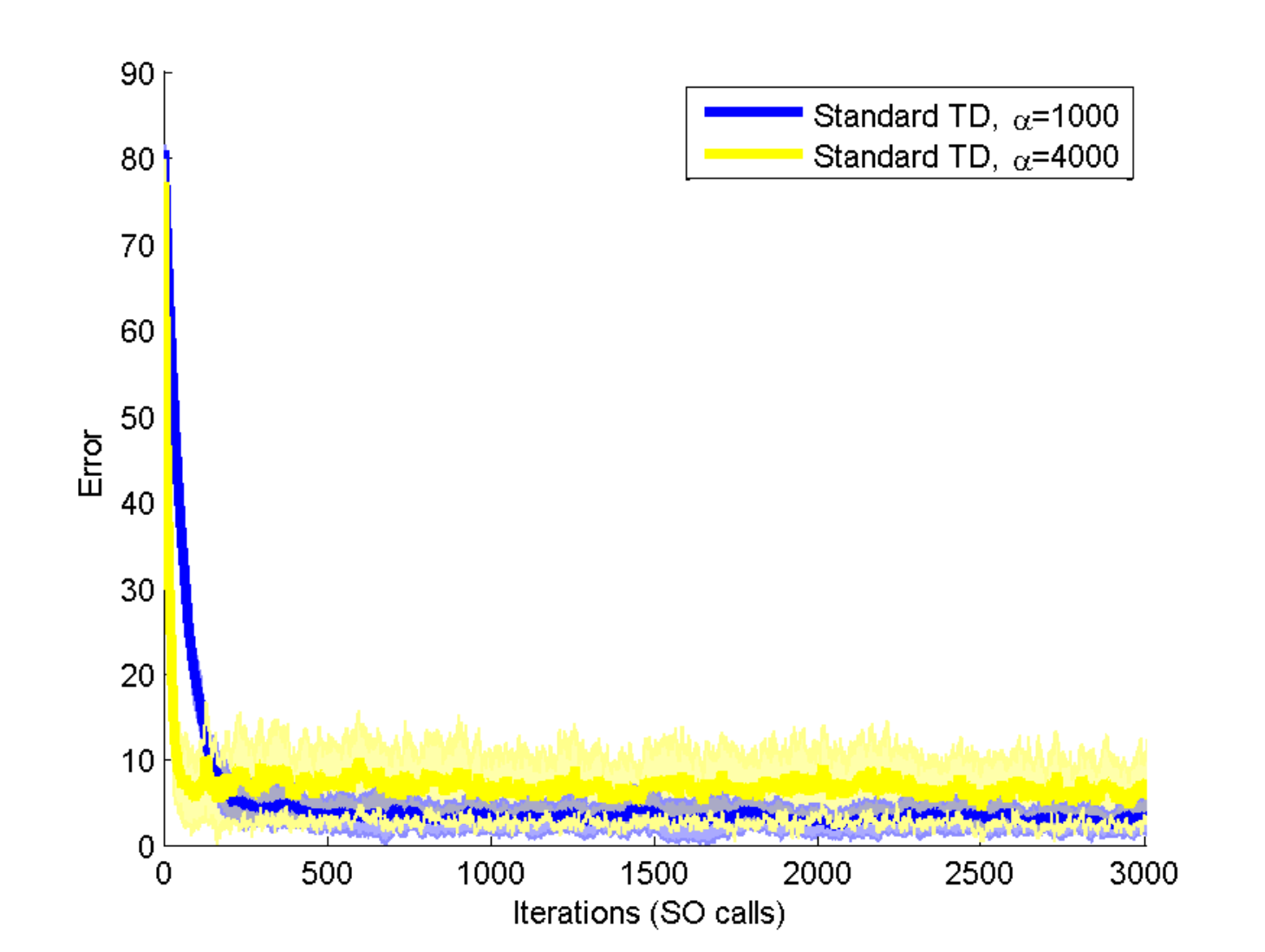}}
\subfigure[{A-TD}]{\includegraphics[width=8cm,height=6cm]{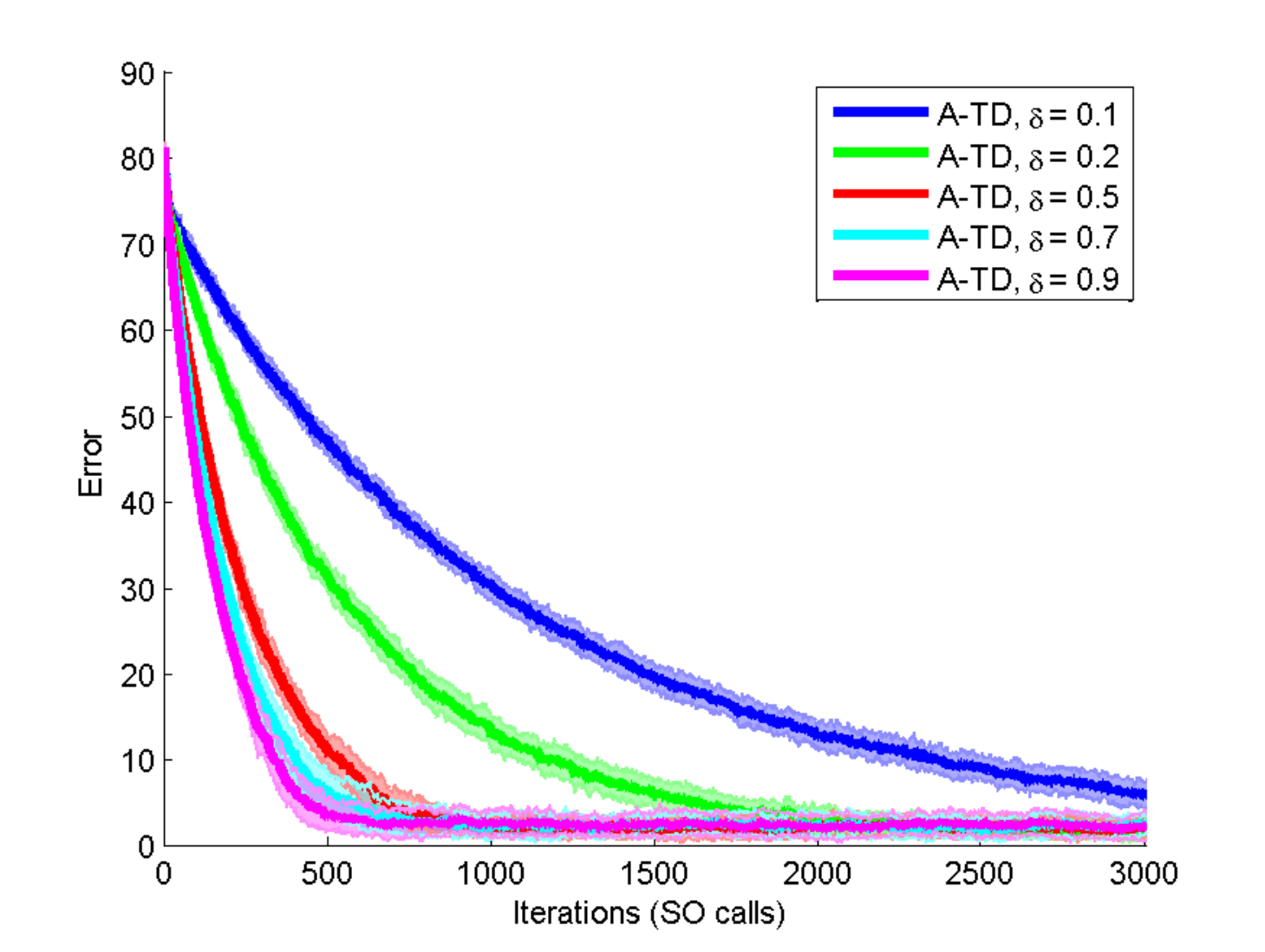}}
\caption{(a) Errors of the standard TD-learning with different step-sizes, $\alpha_k=\alpha/(k+10000)$, $\alpha=1000,4000$. (b) Errors of A-TD with step-size $\alpha_k=1000/(k+10000)$ and $\delta=0.1,0.2,0.5,0.7,0.9$. The shaded areas depict empirical variances obtained with several realizations.}\label{fig:ex2-figure1a}
\end{figure*}

We consider the same MDP as in~\cref{section:simulations} with a linear function approximation using the feature vector
\begin{align*}
&\phi(s) = \begin{bmatrix}
   \frac{\exp(-(s-0)^2)}{2\times 10^2}\\
   \frac{\exp(-(s-10)^2)}{2\times 10^2}
\end{bmatrix} \in {\mathbb R}^2.
\end{align*}
\cref{fig:ex2-figure1a}{(a)} depicts the error evolution of the standard TD-learning with different step-sizes, $\alpha_k=\alpha/(k+10000)$, $\alpha=1000,4000$, by which one concludes that the step-size $\alpha_k=1000/(k+10000)$ provides reasonable performance. \cref{fig:ex2-figure1a}{(b)} illustrates the error evolution of A-TD with step-size $\alpha_k=1000/(k+10000)$ and $\delta=0.1,0.2,0.5,0.7,0.9$. From~\cref{fig:ex2-figure1a}{(b)}, we can observe that the smaller the $\delta$, the slower the convergence rate.

Next, we consider the same MDP as in~\cref{section:simulations} with a linear function approximation using the feature vector
\begin{align*}
&\phi(s) = \begin{bmatrix}
   \frac{\exp(-(s-0)^2)}{2\times 10^2}\\
   \frac{\exp(-(s-10)^2)}{2\times 10^2}\\
   \frac{\exp(-(s-20)^2)}{2\times 10^2}\\
\end{bmatrix} \in {\mathbb R}^3.
\end{align*}

\begin{figure*}[h!]
\centering\subfigure[{Standard TD}]{\includegraphics[width=8cm,height=6cm]{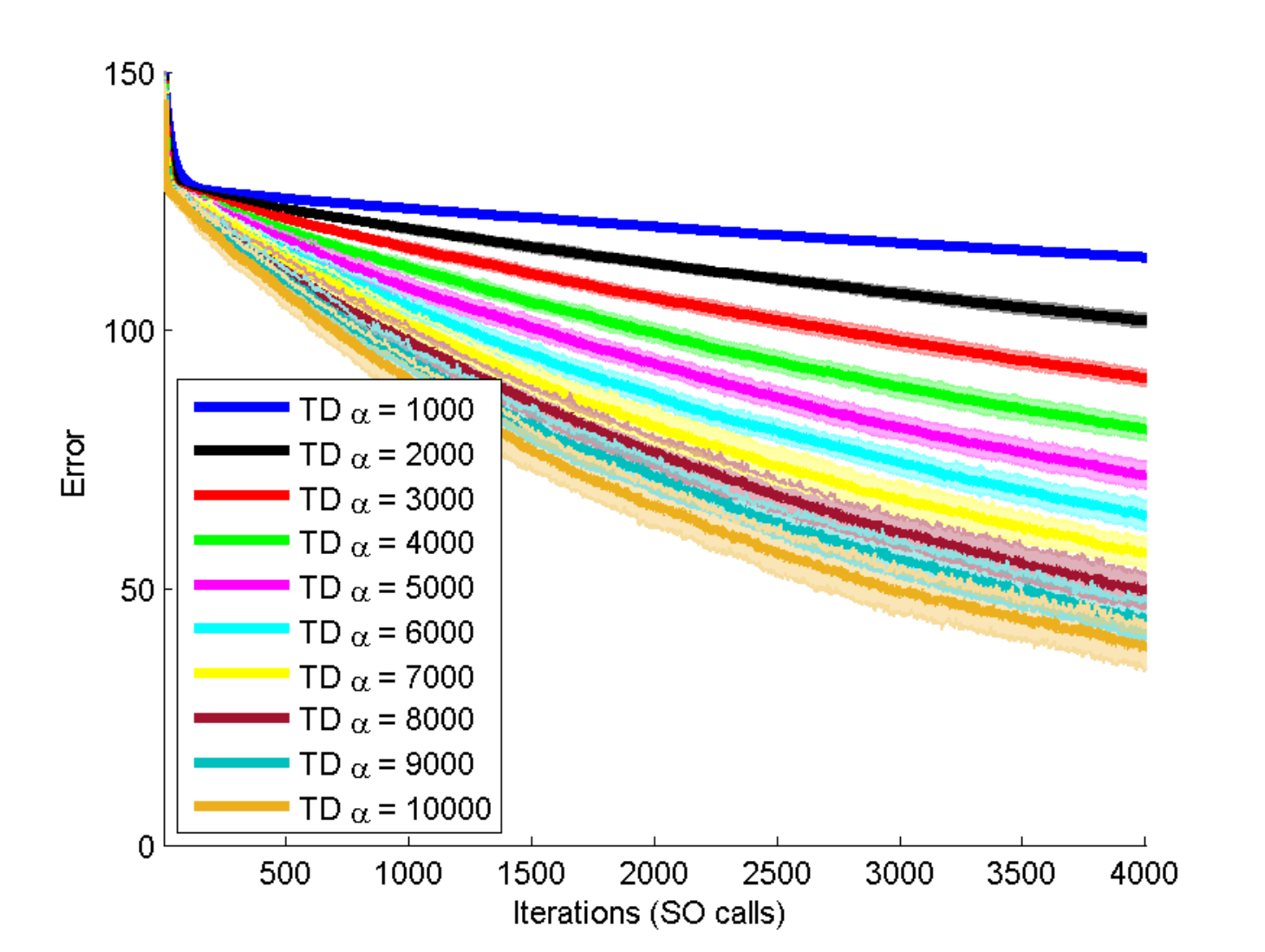}}\subfigure[{Periodic TD, $\beta_t= 4000/(10000 + t)$}]{\includegraphics[width=8cm,height=6cm]{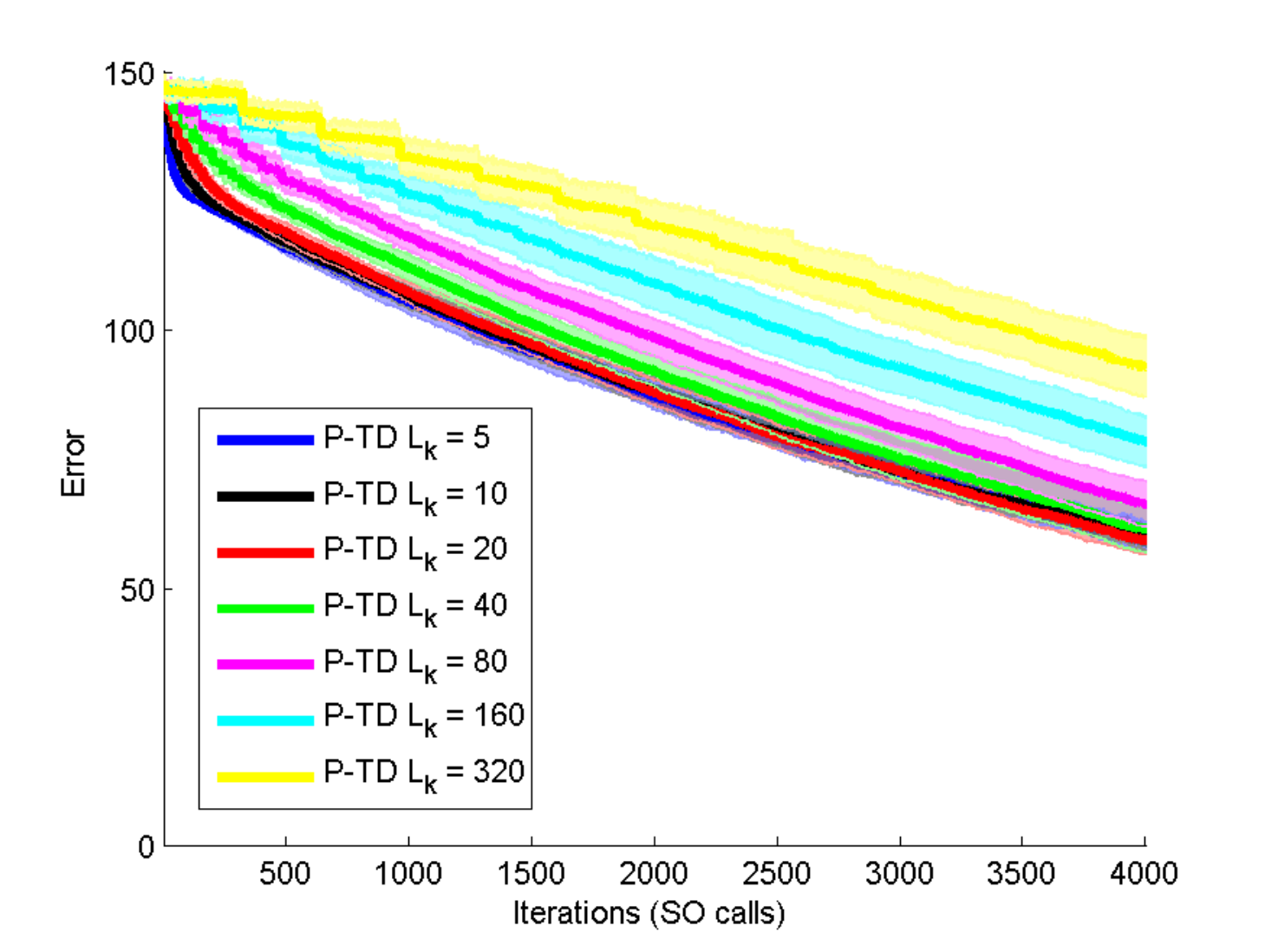}}
\subfigure[{Periodic TD, $\beta_t= 6000/(10000 + t)$}]{\includegraphics[width=8cm,height=5cm]{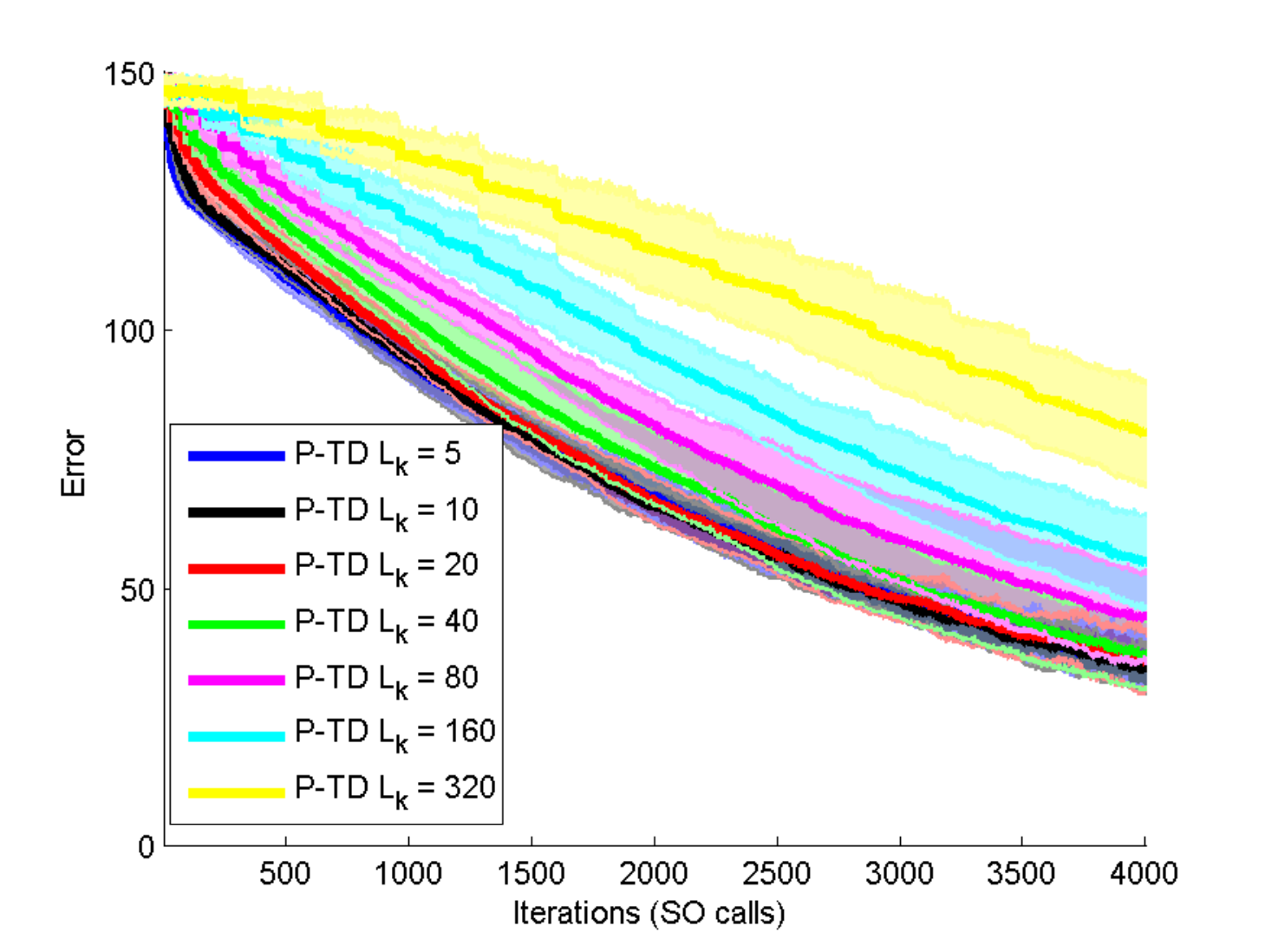}}
\subfigure[{Periodic TD, $\beta_t= 8000/(10000 + t)$}]{\includegraphics[width=8cm,height=5cm]{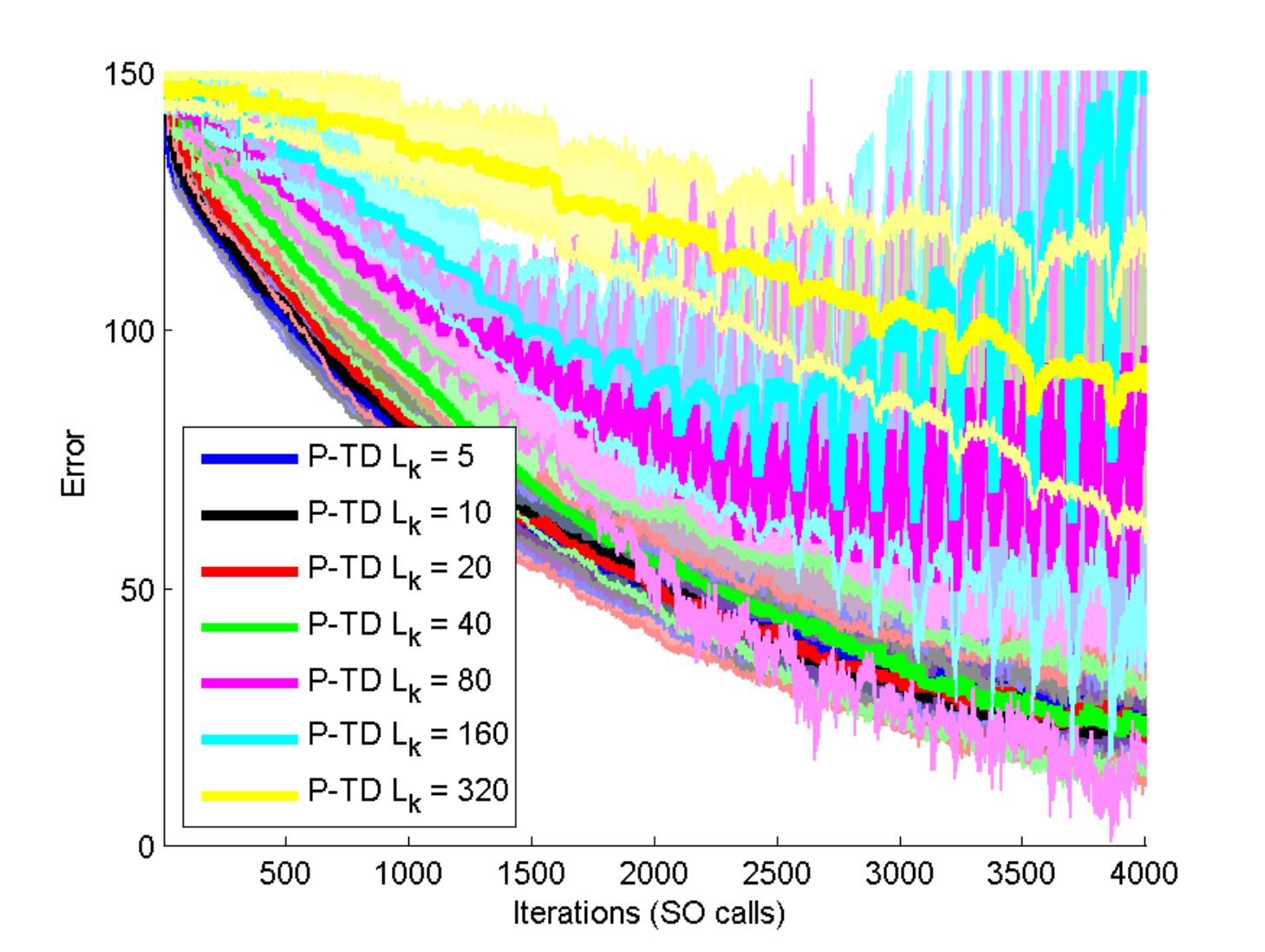}}
\caption{(a) Error evolution of the standard TD-learning with different step-sizes, $\alpha_k=\alpha/(k+10000)$, $\alpha=1000,2000,\dots,10000$. Error evolution of P-TD with $L_k=5,10,20,40,80,160,320$ and the step-sizes, (b) $\beta_t= 4000/(10000 + t)$, (c) $\beta_t= 6000/(10000 + t)$, (d)  $\beta_t= 8000/(10000 + t)$. The shaded areas depict empirical variances obtained with several realizations.}\label{fig:ex2-figure1b}
\end{figure*}

Simulation results (error evolution) for the standard TD are given in~\cref{fig:ex2-figure1b}{(a)} with different step-sizes $\alpha_k=\alpha/(10000+k)$ and $\alpha=1000,2000,\ldots,10000$. Moreover, simulation results of P-TD are given in~\cref{fig:ex2-figure1b}{(b)} with $L_k = 5,10,20,40,80,160,320$, where the following different step-sizes are used: $\beta_t= 4000/(10000+t)$ in~\cref{fig:ex2-figure1b}{(b)}, $\beta_t= 6000/(10000+t)$ in~\cref{fig:ex2-figure1b}{(c)}, $\beta_t= 8000/(10000+t)$ in~\cref{fig:ex2-figure1b}{(d)}.

\begin{figure*}[h!]
\centering\subfigure[{Periodic TD, $L_k=10$}]{\includegraphics[width=8cm,height=6cm]{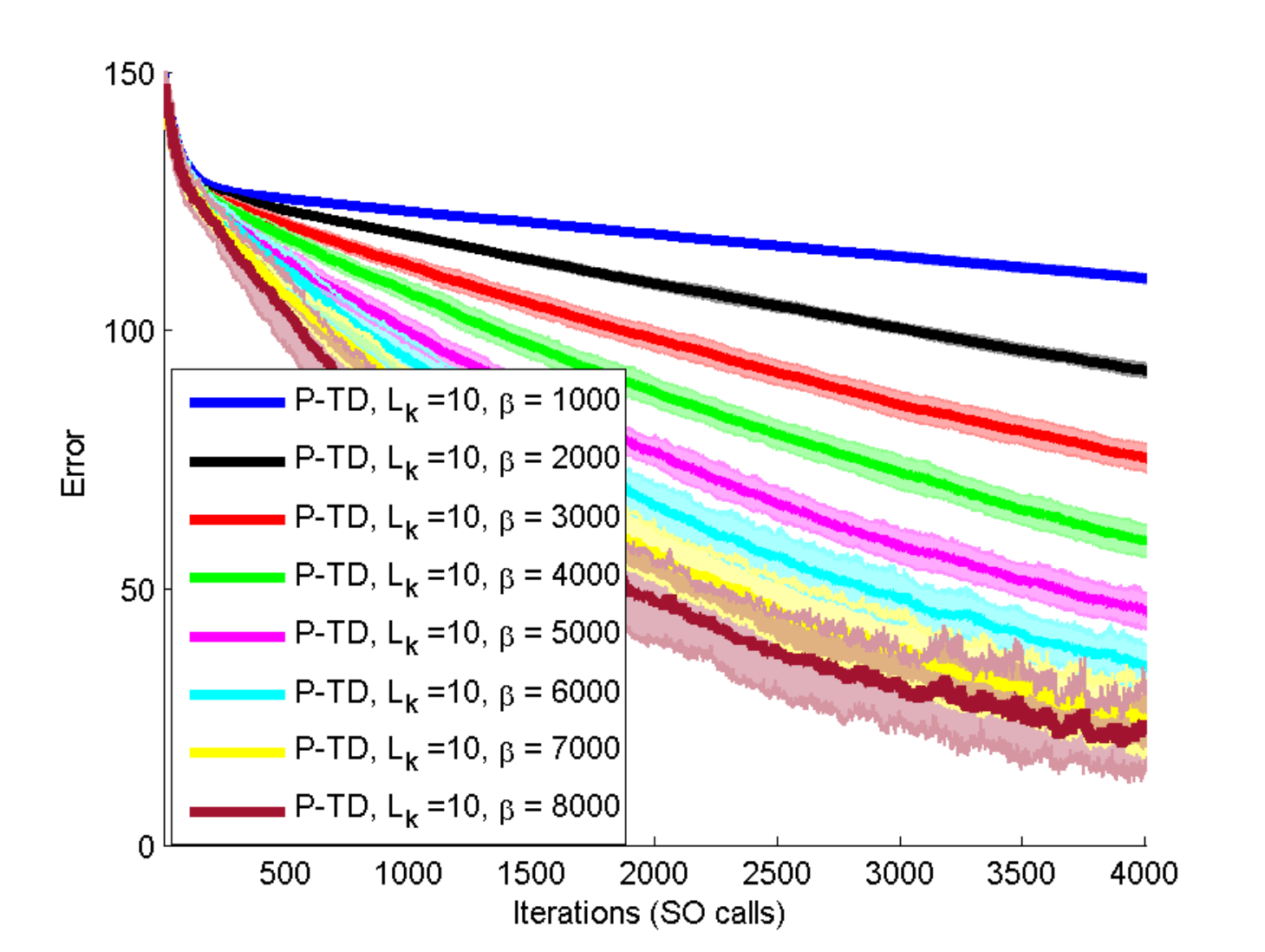}}\subfigure[{Periodic TD, $L_k=20$}]{\includegraphics[width=8cm,height=6cm]{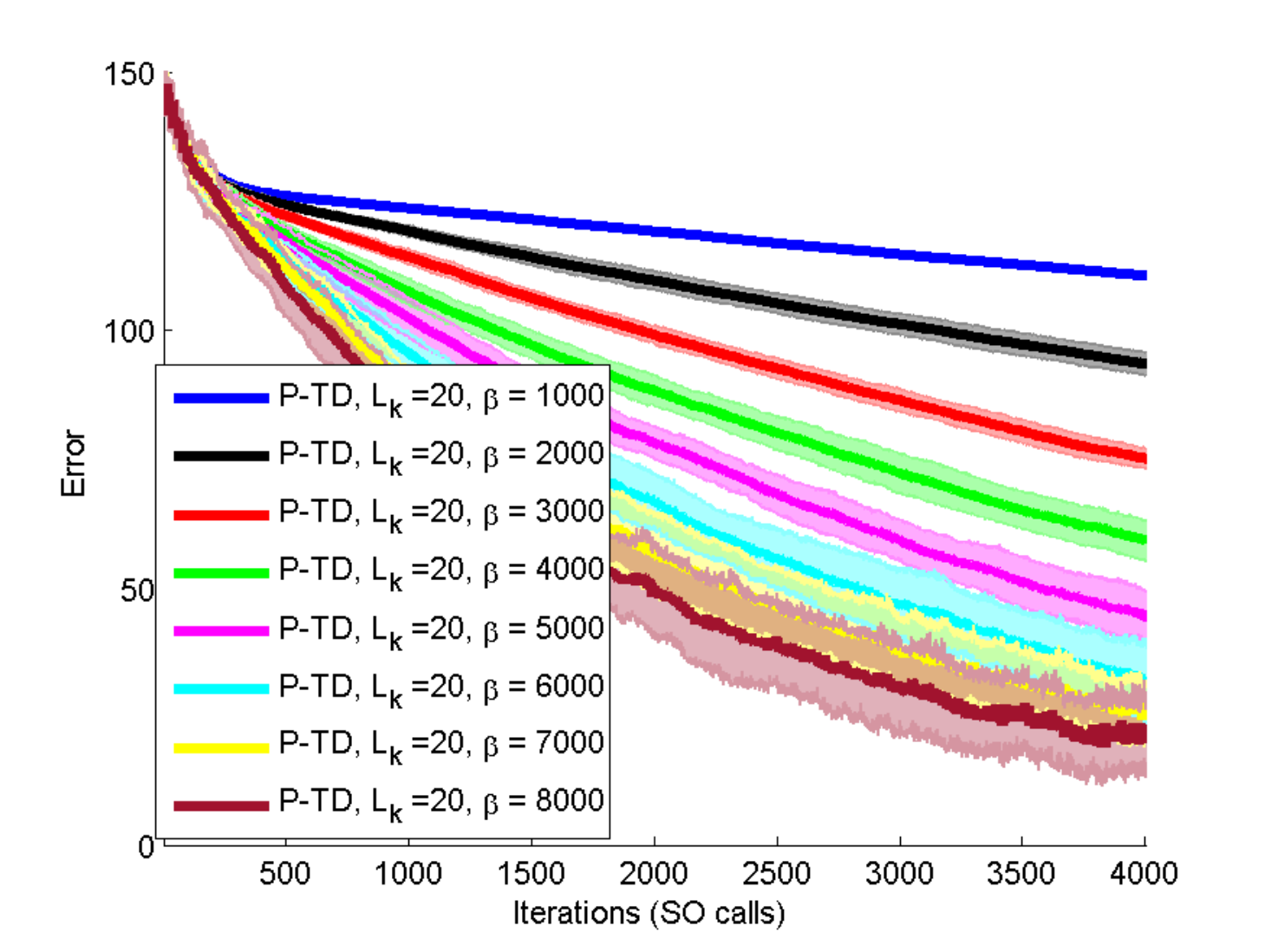}}\\\subfigure[{Periodic TD, $L_k=40$}]{\includegraphics[width=8cm,height=6cm]{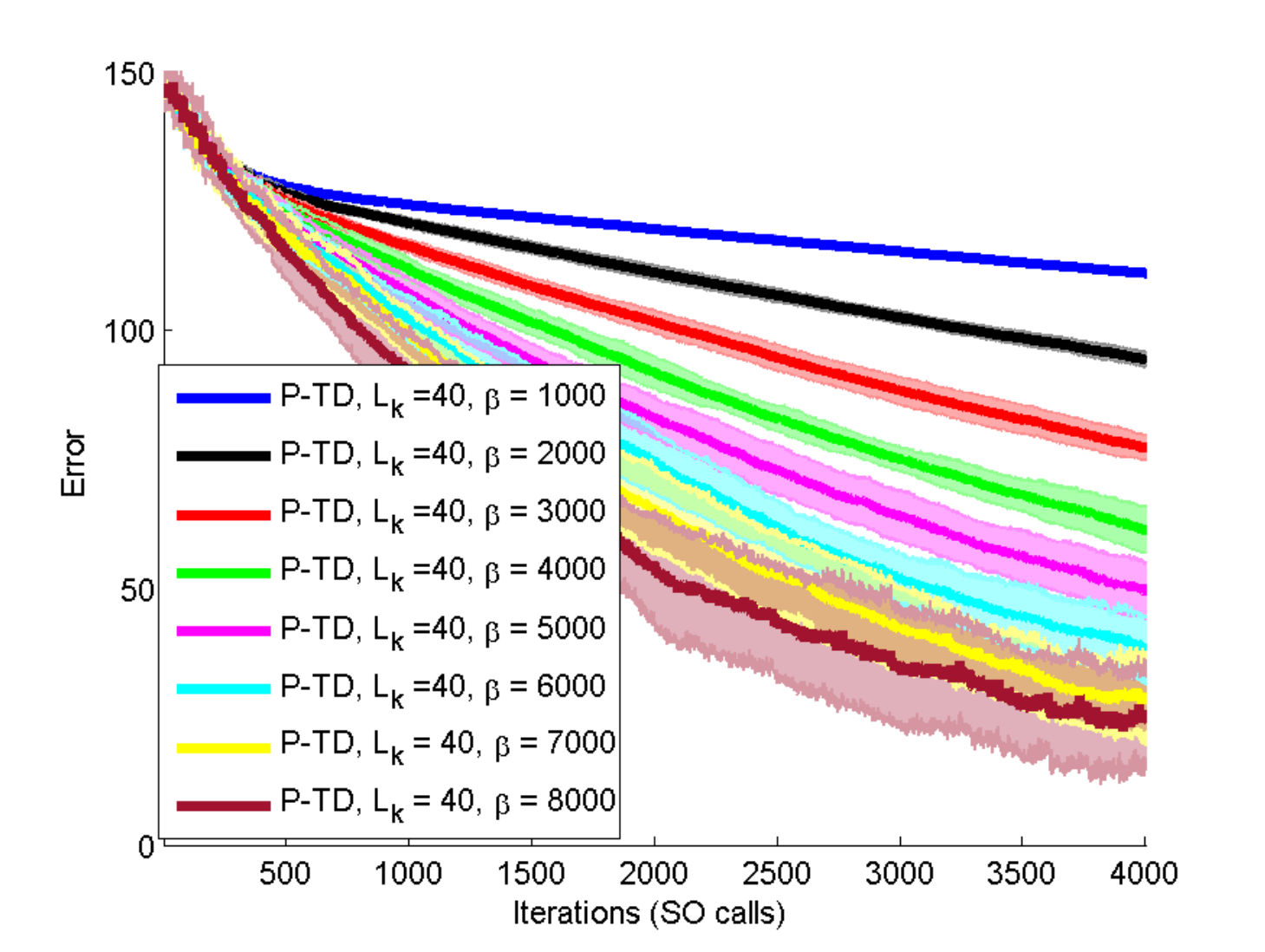}}
\caption{Error evolution of P-TD with different step-sizes, $\beta_t= \beta/(10000+t)$, $\beta=1000,2000,\ldots,8000$. Each subplot uses different $L_k$: (a) $L_k = 10$; (b) $L_k = 20$; (c) $L_k = 40$. The shaded areas depict empirical variances obtained with several realizations. }\label{fig:ex2-figure1c}
\end{figure*}

\cref{fig:ex2-figure1c} illustrates the error plots of P-TD for step-sizes, $\beta_t= \beta/(10000+t)$, $\beta=1000,2000,\ldots,8000$ and different $L_k=10$ (\cref{fig:ex2-figure1c}{(a)}), $L_k=20$ (\cref{fig:ex2-figure1c}{(b)}), and $40$ (\cref{fig:ex2-figure1c}{(c)}).

From~\cref{fig:ex2-figure1a} and~\cref{fig:ex2-figure1c}, one observes that the error evolution for $\beta=8000$ has large fluctuations.

\end{appendix}

\end{document}